\pdfoutput=1

\documentclass[11pt]{article}

\usepackage[usenames,dvipsnames]{xcolor}
\usepackage{graphicx}
\usepackage[ruled,vlined,linesnumbered]{algorithm2e}
\usepackage{todonotes}
\usepackage{CJKutf8}
\usepackage{multirow}
\usepackage{amsmath}
\usepackage{amsthm}
\usepackage{mathtools}
\usepackage{bm}
\usepackage{amsfonts}
\usepackage{booktabs}
\usepackage{subcaption}
\usepackage{lettrine}
\usepackage{placeins}
\usepackage{pifont}
\usepackage{makecell}
\newcommand{\cmark}{\ding{51}}%
\newcommand{\xmark}{\ding{55}}%

\usepackage{acl}

\usepackage{times}
\usepackage{latexsym}

\usepackage[T1]{fontenc}

\usepackage[utf8]{inputenc}

\usepackage{microtype}

\usepackage{etoolbox}

\newcounter{savefootnote}\newcounter{savempfootnote}
\newcounter{symfootnote}\newcounter{symmpfootnote}
\AtBeginEnvironment{minipage}{\setcounter{symmpfootnote}{0}}
\newcommand{\symfootnote}[1]{%
   \setcounter{savefootnote}{\value{footnote}}\setcounter{savempfootnote}{\value{mpfootnote}}%
   \setcounter{footnote}{\value{symfootnote}}\setcounter{mpfootnote}{\value{symmpfootnote}}%
   \ifnum\value{footnote}>8\setcounter{footnote}{0}\fi\ifnum\value{mpfootnote}>8\setcounter{mpfootnote}{0}\fi%
   \let\oldthefootnote=\thefootnote\let\oldthempfootnote=\thempfootnote%
   \renewcommand{\thefootnote}{\fnsymbol{footnote}}\renewcommand{\thempfootnote}{\fnsymbol{mpfootnote}}%
   \footnote{#1}%
   \let\thefootnote=\oldthefootnote\let\thempfootnote=\oldthempfootnote%
   \setcounter{symfootnote}{\value{footnote}}\setcounter{symmpfootnote}{\value{mpfootnote}}%
   \setcounter{footnote}{\value{savefootnote}}\setcounter{mpfootnote}{\value{savempfootnote}}%
}

\usetikzlibrary{shapes.misc, positioning}

\makeatletter
\newcommand{\removelatexerror}{\let\@latex@error\@gobble}
\makeatother

\usepackage[russian, greek, english]{babel}

\SetCommentSty{mycomfont}
\SetKw{Break}{break}

\newtheorem{obs}{Observation}
\newtheorem{cor}{Corollary}

\newcommand{\R}{\mathbb{R}}
\newcommand{\T}{^\top}

\newcommand{\vvec}[1]{\mathbf{#1}}
\newcommand{\hh}[1]{\textbf{#1}}
\newcommand{\norm}[1]{\left\lVert#1\right\rVert}
\newcommand{\perm}[1]{\bm{\pi}_{#1}}
\newcommand{\preg}[1]{\mathcal{R}_{\bm{\pi}_{#1}}}

\DeclareMathOperator*{\argmax}{\arg\!\max}
\DeclareMathOperator*{\row}{row}
\newcommand{\GrantNo}{825303}
\newcommand{\ProjectName}{Bergamot}
\newcommand{\ProjectType}{Research and Innovation Action}

\title{Low-Rank Softmax Can Have Unargmaxable Classes in Theory\\ but Rarely in Practice}

\author{Andreas Grivas \and Nikolay Bogoychev \and Adam Lopez \\
Institute for Language, Cognition, and Computation\\
School of Informatics\\
University of Edinburgh\\
  \texttt{\{agrivas, n.bogoych, alopez\}@ed.ac.uk} 
  }

\begin{document}

\maketitle
\begin{abstract}
Classifiers in natural language processing (NLP) often have a large number of output classes. For example, neural language models (LMs) and machine translation (MT) models both predict tokens from a vocabulary of thousands. The Softmax output layer of these models typically receives as input a dense feature representation, which has much lower dimensionality than the output. In theory, the result is some words may be impossible to be predicted via argmax, irrespective of input features, and empirically, there is evidence this happens in small language models \citep{demeter2020}. In this paper we ask whether it can happen in practical large language models and translation models. To do so, we develop algorithms to detect such \emph{unargmaxable} tokens in public models. We find that 13 out of 150 models do indeed have such tokens; however, they are very infrequent and unlikely to impact model quality. We release our code so that others can inspect their models.\footnote{\label{code1}\href{https://github.com/andreasgrv/unargmaxable}{https://github.com/andreasgrv/unargmaxable}}

\end{abstract}

\section{Introduction}

Probabilistic multiclass classifiers with a large number of output classes are commonplace in NLP~\citep{chen2016}.
For example, the vocabulary size of contemporary LMs and MT models varies from tens to hundreds of thousands~\citep{liu-etal-2020}.
Recent advances in modelling such large vocabularies have mostly been made by improving neural network feature encoders~\cite{devlin2018,conneau-etal-2020-roberta}.
But irrespective of a feature encoder's expressivity~\citep{yun2020, raghu2017}, a classifier that linearly maps lower dimensional features to higher dimensional outputs has reduced expressivity~\citep{yang2018}, with consequences that are not well understood.

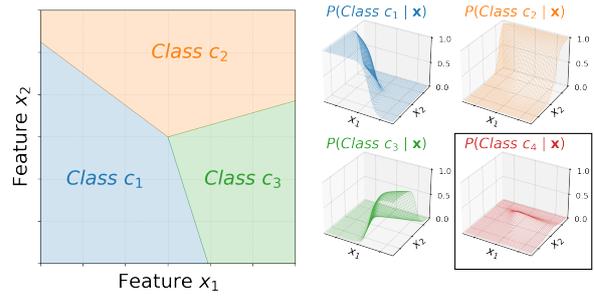
\begin{figure}[t!]
\begin{tikzpicture}[inner sep=0pt, overlay, remember picture] 
\node[rectangle, draw, minimum height=1.8cm, minimum width=1.8cm] at (6.85, 1.2) {};
\end{tikzpicture}
\includegraphics[width=\columnwidth]{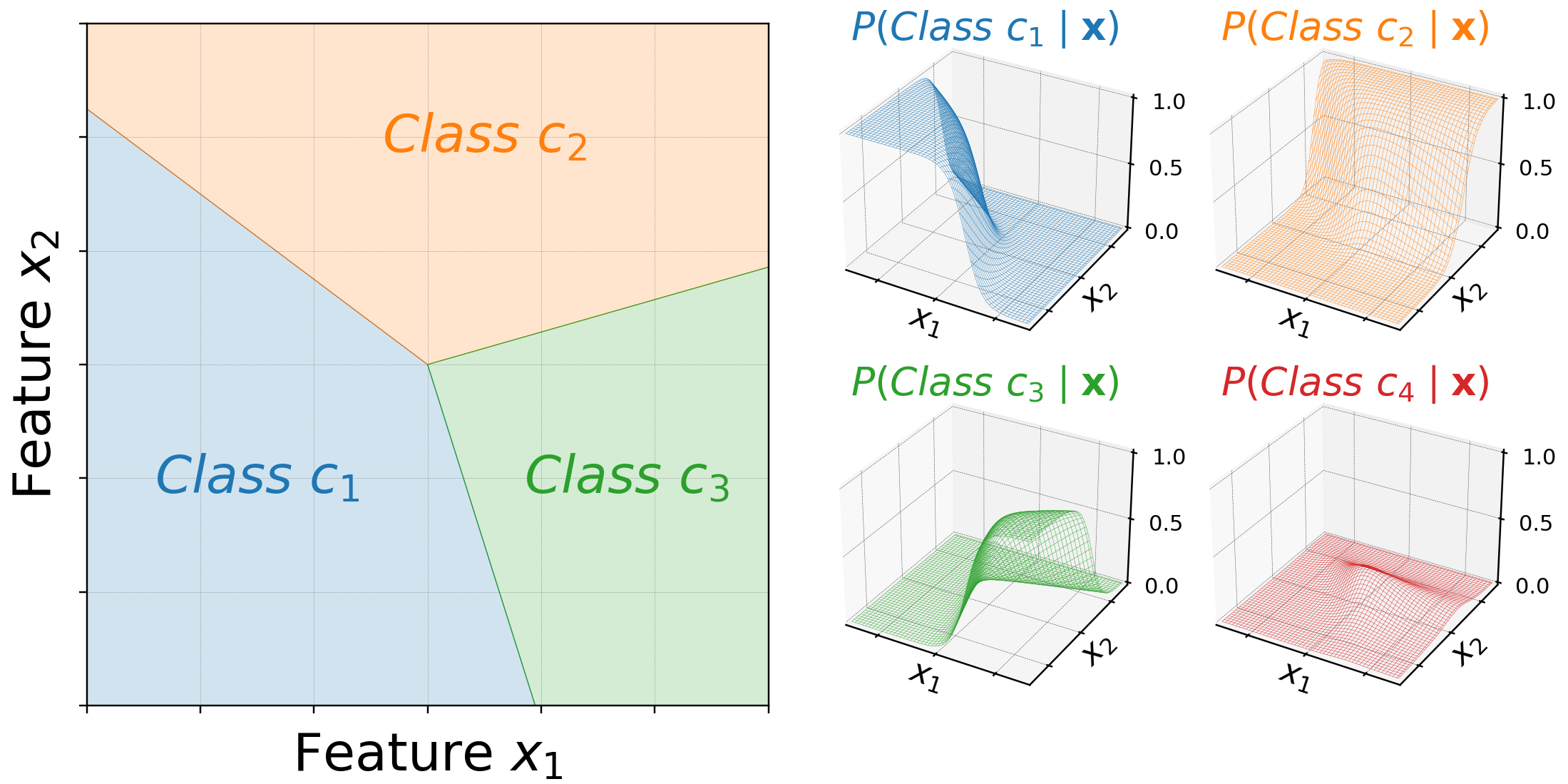}
\caption{Illustration of an \emph{unargmaxable} class. \textcolor{BrickRed!60!red}{Class $c_4$} can never be predicted using argmax for this Softmax classifier with $|C| = 4$ classes and $d=2$ input features. On the left, each feature vector $\vvec{x}$ is colored according to the class assigned the largest probability; note that while \textcolor{RoyalBlue}{$c_1$}, \textcolor{Orange}{$c_2$} and \textcolor{Green}{$c_3$} surface as regions, \textcolor{BrickRed!60!red}{$c_4$} does not. On the right, we show that there is no direction in feature space for which \textcolor{BrickRed!60!red}{$c_4$} has the largest probability.
 }
\label{fig:stolen}
\end{figure}

In this work we elaborate on the consequences of using argmax prediction with low-rank classifiers, classifiers that have more output classes $|C|$ than features $d$. 
For example, MT models often have subword vocabularies of size $|C| \approx 30000$, but have $d \approx 1024$.
The expressivity penalty for such low-rank classifiers is that some output distributions cannot be represented.
\citet{demeter2020} identified this weakness in Softmax LMs, showing that, in theory, some tokens can never be assigned the highest probability for any input, and therefore can never be produced as argmax predictions.\footnote{This problem was also studied by~\citet{Cover1967} and has an interesting history of independent discovery~\citep{Smith2014}.} We call such tokens \textbf{unargmaxable} (see Figure~\ref{fig:stolen}).

While \citet{demeter2020} proposed an algorithm to detect unargmaxable tokens and provided evidence of their existence in small LMs, their proposed algorithm provided no guarantees and they were unable to test large LMs. In this paper we ask: \textit{Do unargmaxable tokens exist in large models used in practice?}
To answer this question, we develop algorithms to identify such tokens unambiguously. We tested $7$ LMs and $143$ MT models. Out of those, only $13$ of the MT models exhibit unargmaxable tokens, and even for those cases the tokens are all noisy and infrequent. We conclude that although the expressivity constraints of low-rank Softmax may have important ramifications, most practitioners do not need to worry about tokens that are unargmaxable. We provide new tools for them to confirm this on their own models.

Our contributions are the following:

\begin{itemize}
    \setlength\itemsep{.1em}
    \item We explain how unargmaxable tokens can arise as a consequence of a rank constrained Softmax layer (Softmax Bottleneck).
    \item We extend the work of \citet{demeter2020} with verification algorithms that include the Softmax bias term and provide an exact answer rather than an approximate one.
    \item We verify a large number of commonly used publicly available language and translation models for unargmaxable tokens.
    \item We release our algorithm and code so that others can inspect their models.\textsuperscript{\ref{code1}}
\end{itemize}

\section{Background}

\subsection{Low-Rank Softmax (Softmax Bottleneck)}

Neural network layers with higher dimensional outputs than inputs impose low-rank constraints.\footnote{A layer can also be low rank if weight vectors are collinear, but we do not consider this case here.}
Such constraints commonly exist as \textbf{bottlenecks} in neural network hidden layers, e.g. autoencoders~\cite{Hinton1993} and projection heads in multi-head transformers~\cite{bhojanapalli20a} among others.
While bottlenecks make a model less expressive by restricting the functions it can represent, they are desirable both computationally~\citep{papadimitriou2021}, since they require less memory and computation than full-rank layers, and as a form of inductive bias, since data is assumed to approximately lie in a low dimensional manifold~\citep{mcinnes2018umap}.

In contrast, herein we focus on the undesirable properties of a Softmax output layer with a low-rank parametrisation, also known as a ~\textbf{Softmax Bottleneck}~\citep{yang2018}.
The crucial difference is that a Softmax Bottleneck is usually not followed by a non-linear transformation, and as such the rank constraint limits expressivity in a very rigid way by restricting outputs to a subspace.\footnote{A linear subspace if no bias term is present and an affine subspace otherwise.}
This constraint was shown to hurt LM perplexity~\cite{yang2018} and non-linear augmentations have been proposed as improvements~\citep{yang2018, Kanai2018, Ganea2019}. 
To the contrary, \citet{sainath2013} used a low-rank factorisation of the softmax layer to reduce the number of parameters in their speech recognition system by $30$-$50\%$ with no increase in word-error-rate, evidencing that the loss in expressivity does not always impact aggregate metrics.

The consequences of the loss in expressivity due to the Softmax Bottleneck vary depending on our perspective.
When considering the flexibility of the probability distribution that can be learned, \citet[Theorem 2]{Ganea2019} showed that the minimum cross entropy loss achievable decreases as we increase the rank of the Softmax layer weights.

In this work we focus on the loss of expressivity from an argmax perspective.
To this end, we discretise the output space of Softmax and quantify the loss in expressivity in terms of unrealisable class rankings.
From this interpretable perspective we will see that due to the Softmax Bottleneck some rankings are not realisable and unargmaxable classes can arise as a consequence.

\subsection{Unargmaxable Classes}

\begin{figure}[t!]
        \includegraphics[width=\columnwidth]{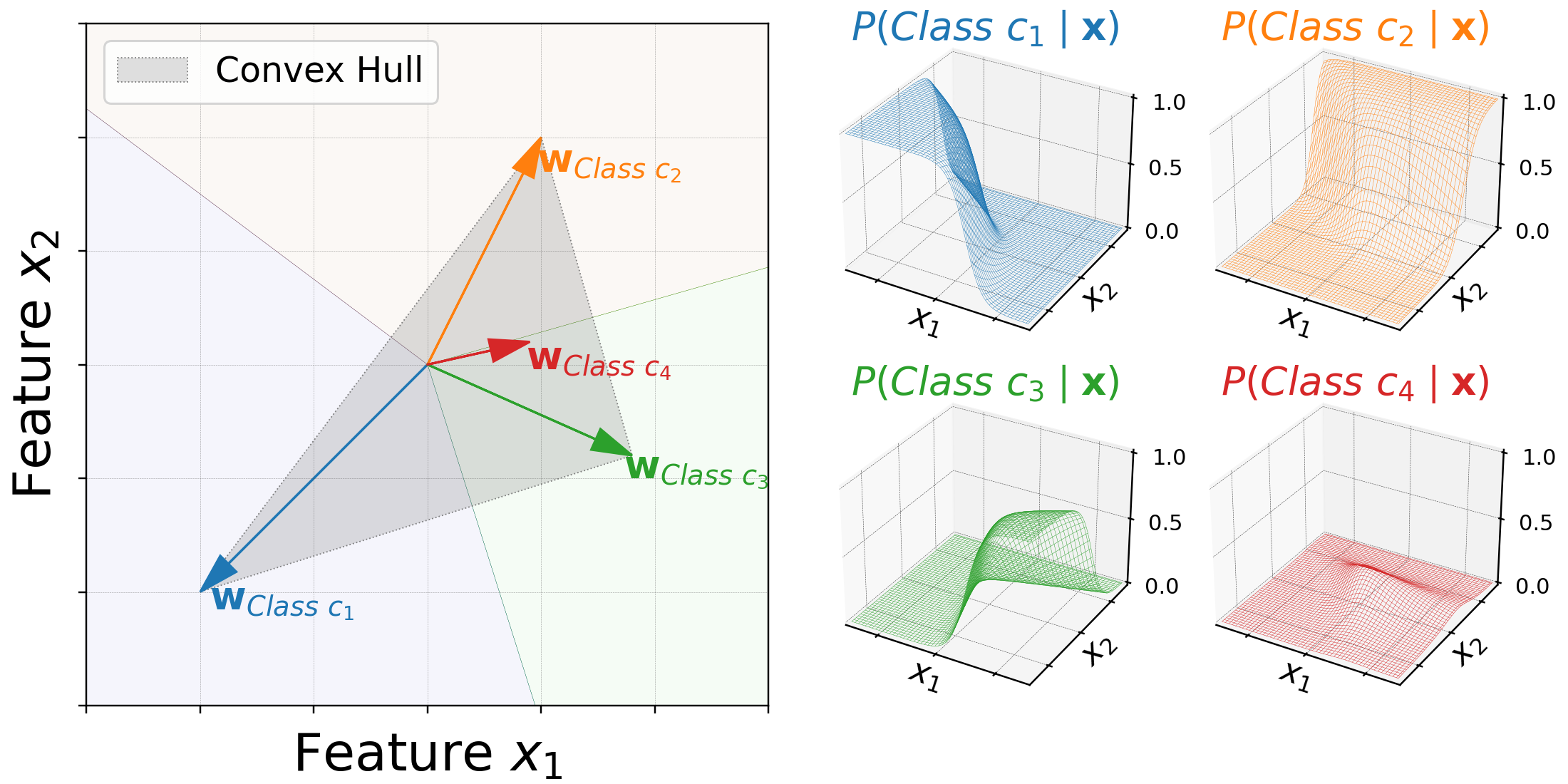}
\caption{Illustration of how \emph{unargmaxable} classes arise. The vectors on the left are the culprit Softmax weights for Figure~\ref{fig:stolen}. Each vector is a row of the Softmax weights $\mathbf{W} \in \R^{4 \times 2}$. \textcolor{BrickRed!60!red}{$c_4$} is interior to the convex hull, the triangle formed by \textcolor{RoyalBlue}{$c_1$}, \textcolor{Orange}{$c_2$} and \textcolor{Green}{$c_3$}.}
\label{fig:stolen2}
\end{figure}

\citet{demeter2020} showed that a class is unargmaxable if its Softmax weight vector is interior to the convex hull of the remaining class weight vectors. They did so by proving that the interior class probability is bounded above by the probability of at least one class on the convex hull (see Figure~\ref{fig:stolen2} and \citealp[Figure 1]{Cover1967}). However, in their analysis they did not address Softmax layers that include a bias term. We address this limitation in Section~\ref{sec:detect}, thus enabling us to search for unargmaxable classes in any released model.

To detect whether unargmaxable tokens arise in LMs without a bias term, the authors introduce an approximate algorithm that asserts whether a weight vector is internal to the convex hull.
It is approximate since their method had a precision approaching 100\% but 68\% recall when compared to an exact algorithm (Qhull,~\citealp{barber1996}) on the first $10$ dimensions of a Softmax LM. In Section~\ref{sec:exact} we introduce an exact algorithm to detect unargmaxable tokens with certainty.

The authors use their approximate algorithm to show that AWD-LSTM LMs~\citep{merity2018} ``steal'' probability from candidate interior words when contrasted to the probabilities assigned by a smoothed n-gram LM.
However, they find that as they increase the dimensionality $d$ of the Softmax weights to $200$, the effect of stolen probability begins to dissipate.
This raises the question of whether stolen probability is of importance for neural models used in practice which also have larger Softmax weight dimensionality.

Herein we specifically search for unargmaxable tokens in MT and LM models with larger $d \in [256, 512, 1024]$.
We use the term unargmaxable rather than stolen probability to highlight that we are focussing on whether unargmaxable tokens exist and not whether the probability distibution learned by low-rank Softmax is less flexible.
We extend our analysis to MT models since they have more practical use cases than (generative) LMs: if unargmaxable tokens exists in a MT model, then the affected tokens can never be produced when using greedy decoding.
In our experiments we find that while unargmaxable tokens arise in limited cases, they are not of grave importance.

\section{Detecting Unargmaxable Classes}
\label{sec:detect}
In order to quantify whether unargmaxable classes arise in released LMs and MT models, we first need to introduce tractable algorithms for detecting them. In this Section we explain how unargmaxable classes can arise due to a Softmax Bottleneck. Then, we introduce a fast approximate algorithm and a slow exact algorithm which we combine to detect vocabulary tokens that cannot be predicted.

\subsection{Definitions}
We use boldface for matrices and vectors. All vectors are column vectors.
We use $\vvec{w}_i$ for the $i$th row of $\vvec{W}$
and $b_i$ for the $i$th element of $\vvec{b}$.

\subsubsection{Softmax}
A Softmax layer gives us the probability assigned to a target class $c_t$ for an input feature vector $\mathbf{x} \in \R^{d}$ as follows:
\begin{align}
    P(C=c_t \mid \mathbf{x}) &=  \frac{e^{\mathbf{w}_{c_t}\T \mathbf{x} + b_{c_t}}}{\sum_{i} e^{\mathbf{w}_{c_i}\T \mathbf{x} + b_{c_i}}} \\
     &= \operatorname{softmax}(\mathbf{Wx} + \mathbf{b})_{c_t}
\end{align}
where $\mathbf{W} \in \R^{|C| \times d}$ are the class weight vectors stacked row by row, and $\mathbf{b} \in \R^{|C|}$ is the bias term. The above are used to compute the \textbf{logits} $\vvec{y} = \vvec{W}\vvec{x} + \vvec{b}$. In what follows, we will refer to the feature activations $\vvec{x}$ in $\R^d$ as the \textbf{input space} and the logits $\vvec{y}$ in $\R^{|C|}$ as the \textbf{output space} of the Softmax layer.

\subsubsection{Discretising the Output Space into Permutations}
As we saw in Figure~\ref{fig:stolen2}, there are certain arrangements of Softmax weights for which a target class $c_t$ cannot be surfaced as the argmax.
To understand this phenomenon, it will be helpful to discretise the outputs to a finer granularity: rankings~\citep{burges2005}.
In order for a classifier to predict a class $c_t$ using an argmax decision rule, it must rank $c_t$ above all other classes by assigning it the largest probability.
From this perspective, a classifier assigns each input $\vvec{x}$ a permutation $\perm{}$ that ranks the class indices in increasing order of probability.
\begin{equation}
\label{eq:ranking}
\resizebox{.99 \columnwidth}{!}{%
$\perm{} : P(c_{\pi_1} \mid \vvec{x}) < P(c_{\pi_2} \mid \vvec{x}) < \ldots < P(c_{\pi_{|C|}} \mid \vvec{x})$%
}
\end{equation}
As an example, if we have 4 classes and obtain probabilities $P(C \mid \vvec{x}) = \begin{bmatrix} \textcolor{RoyalBlue}{.2} & \textcolor{Orange}{.4} & \textcolor{Green}{.1} & \textcolor{BrickRed!60!red}{.3} \end{bmatrix}\T$ we assign $\vvec{x}$ the permutation $\perm{3142}$ , since $P(\textcolor{Green}{c_3} \mid \vvec{x}) < P(\textcolor{RoyalBlue}{c_1} \mid \vvec{x}) < P(\textcolor{BrickRed!60!red}{c_4} \mid \vvec{x}) < P(\textcolor{Orange}{c_2} \mid \vvec{x})$.
We can readily obtain the coarser argmax decision (\textcolor{Orange}{$c_2$}) by reading off the last index of the permutation.

\subsection{How Can Unargmaxable Classes Arise?}

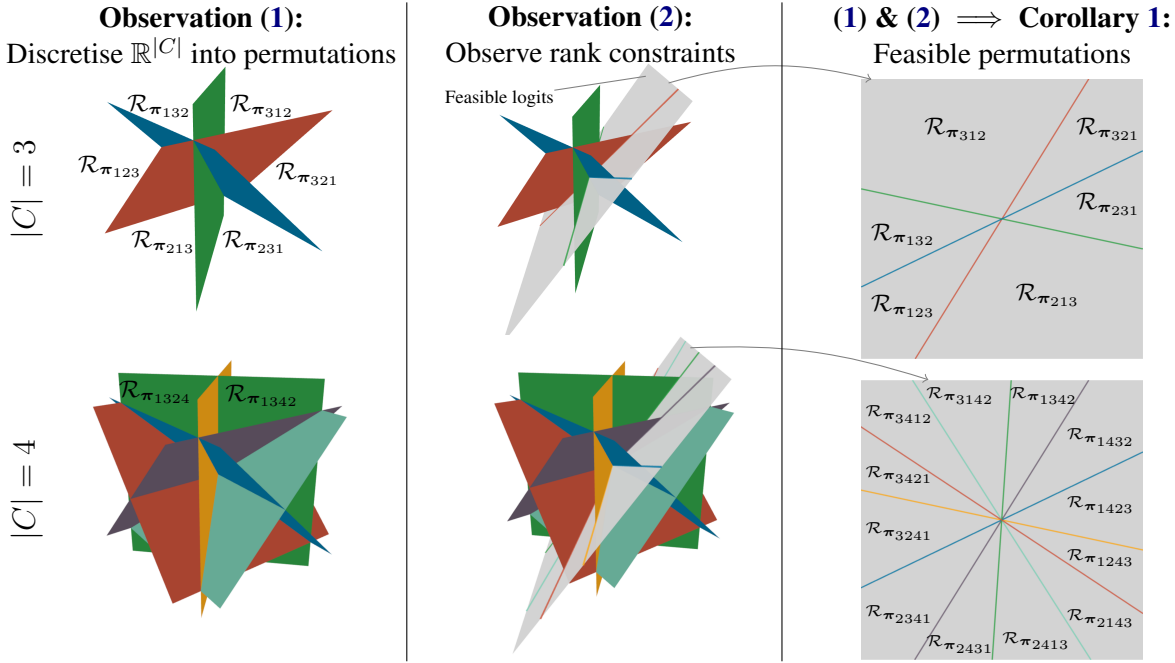
\begin{figure*}[h]
    \vspace{3.5cm}
    \begin{subfigure}[b]{0.32\textwidth}
       \begin{tikzpicture}[inner sep=0pt, overlay, remember picture] 
            \def\x{2.8}
            \def\y{1.}
            \node at (\x, \y) {\includegraphics[height=5cm]{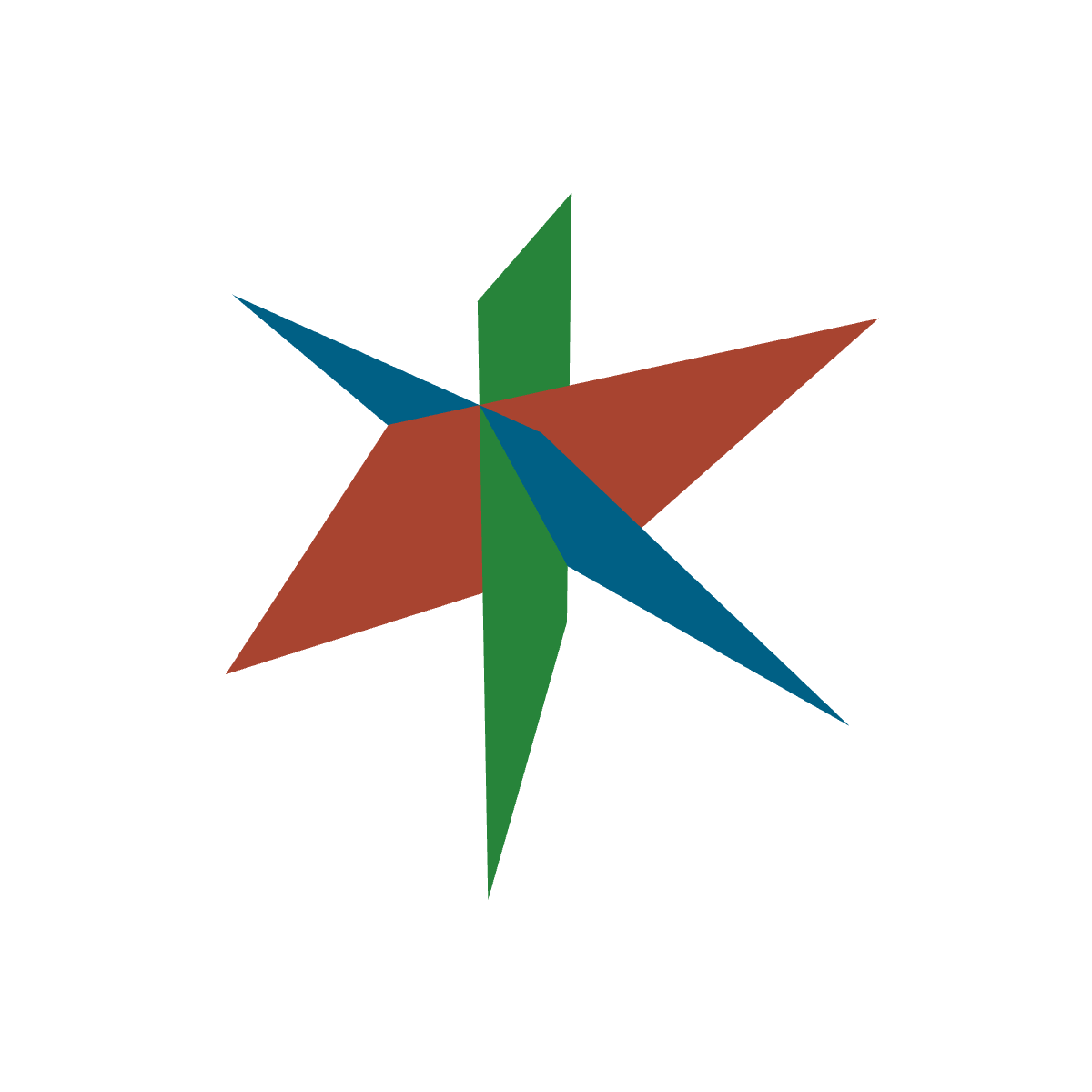}};
            \node[align=center] at (\x -.2, \y + 2) {\textbf{Observation (\ref{obs:part}):}\\ Discretise $\R^{|C|}$ into permutations};
            \node at (\x - 1.4, \y + .3) {\footnotesize{$\preg{123}$}};
            \node at (\x + 1.2, \y + .2) {\footnotesize{$\preg{321}$}};
            \node (pi3) at (\x + .6, \y + 1.1) {\footnotesize{$\preg{312}$}};
            \node at (\x - .75, \y + 1.1) {\footnotesize{$\preg{132}$}};
            \node at (\x - .7, \y - .7) {\footnotesize{$\preg{213}$}};
            \node at (\x + .5, \y - .7) {\footnotesize{$\preg{231}$}};
            \node[align=center, rotate=90] at (.3, \y) {\large{$|C| = 3$}};
            \node (c1) at (\x + 2.5, \y + 2.5) {};
        \end{tikzpicture}
    \end{subfigure}%
    \begin{subfigure}[b]{0.32\textwidth}
       \begin{tikzpicture}[inner sep=0pt, overlay, remember picture] 
            \def\x{2.8}
            \def\y{1.}
            \node at (\x, \y) {\includegraphics[height=4.25cm]{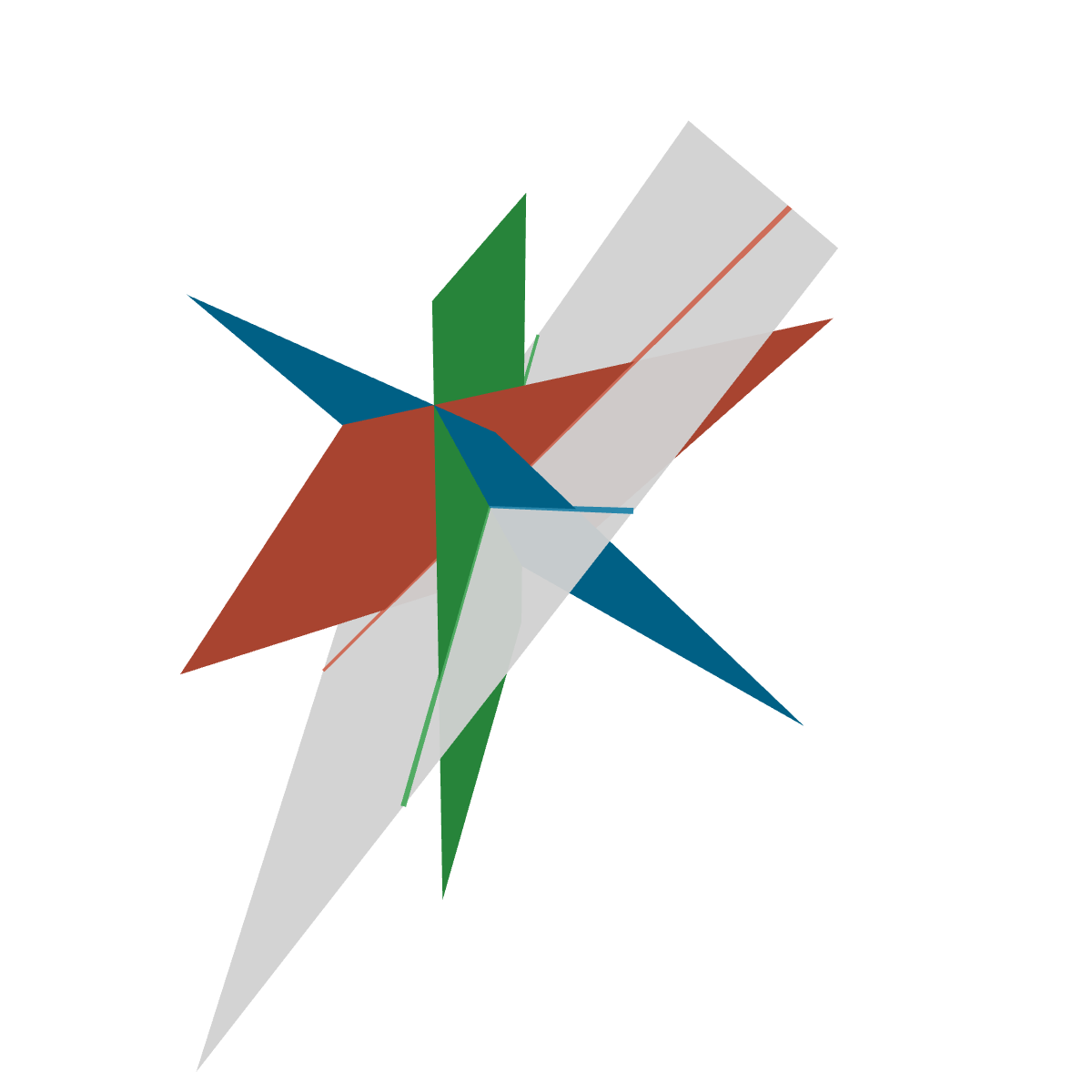}};
            \node[align=center] at (\x - .2, \y + 2.05) {\textbf{Observation (\ref{obs:feas}):}\\ Observe rank constraints};
            \node(p) at (\x + .55, \y + 1.5) {};
            \node(pl) at (\x + .8, \y + 1.4) {};
            \node (note) at (\x - 1.4,  \y + 1.2) {\scriptsize{Feasible logits}};
            \draw[-, color=gray] (note) edge (p);
        \end{tikzpicture}
    \end{subfigure}
    \begin{subfigure}[b]{0.33\textwidth}
       \begin{tikzpicture}[inner sep=0pt, overlay, remember picture] 
            \def\y{.8}
            \node at (2.8, \y - .2) {\includegraphics[height=3.7cm]{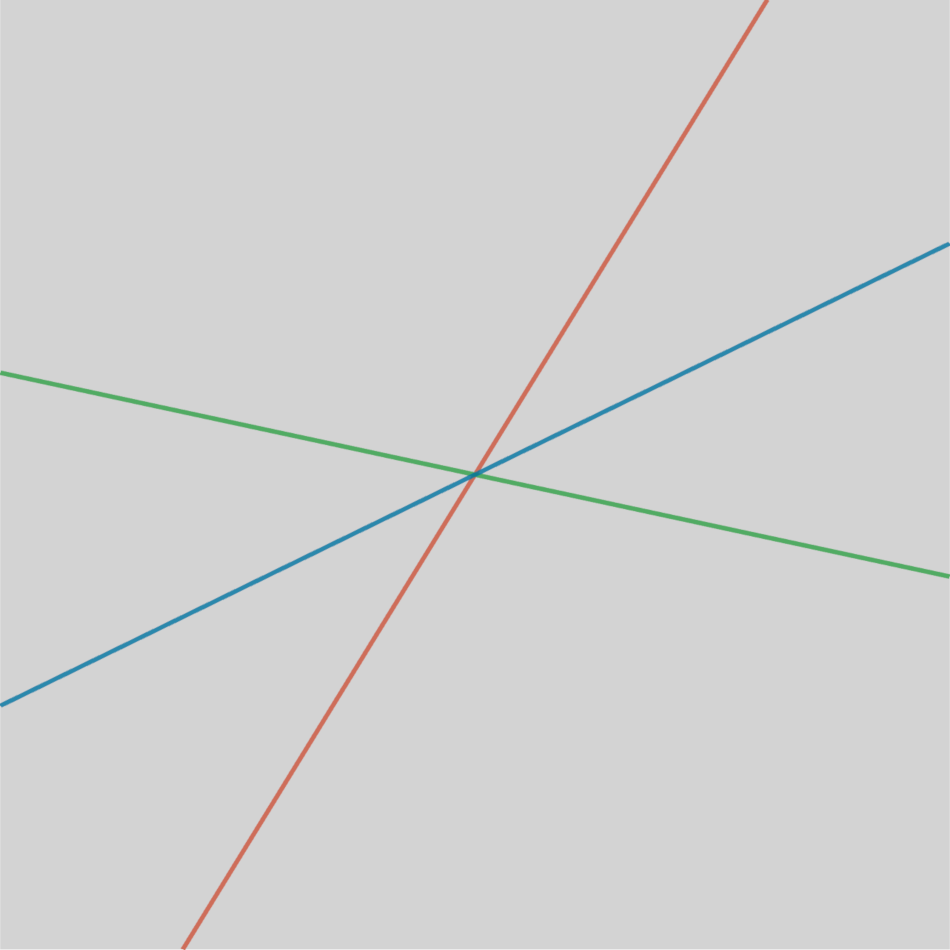}};
            \node[align=center] at (2.8, \y + 2.2) {\textbf{(\ref{obs:part}) \& (\ref{obs:feas}) $\implies$ Corollary \ref{cor:stolen}:}\\ Feasible permutations};
            \node (pra) at (2.2, \y + 1.) {\footnotesize{$\preg{312}$}};
            \node (prb) at (4.2, \y + 1.) {\footnotesize{$\preg{321}$}};
            \node (prc) at (4.2, \y ) {\footnotesize{$\preg{231}$}};
            \node (prd) at (3.4, \y - 1.2) {\footnotesize{$\preg{213}$}};
            \node (pre) at (1.5, \y - 1.35) {\footnotesize{$\preg{123}$}};
            \node (prf) at (1.5, \y - .4) {\footnotesize{$\preg{132}$}};
            \node (pl2) at (1.1, \y + 1.65) {};
            \draw[->, color=gray] (pl) edge [bend left=15] (pl2);
            \node (c3) at (-.1, \y + 2.7) {};
        \end{tikzpicture}
    \end{subfigure}
    
    \begin{subfigure}[b]{0.32\textwidth}
       \begin{tikzpicture}[inner sep=0pt, overlay, remember picture] 
            \def\x{2.8}
            \def\y{2.2}
            \node at (\x, \y) {\includegraphics[height=3.8cm]{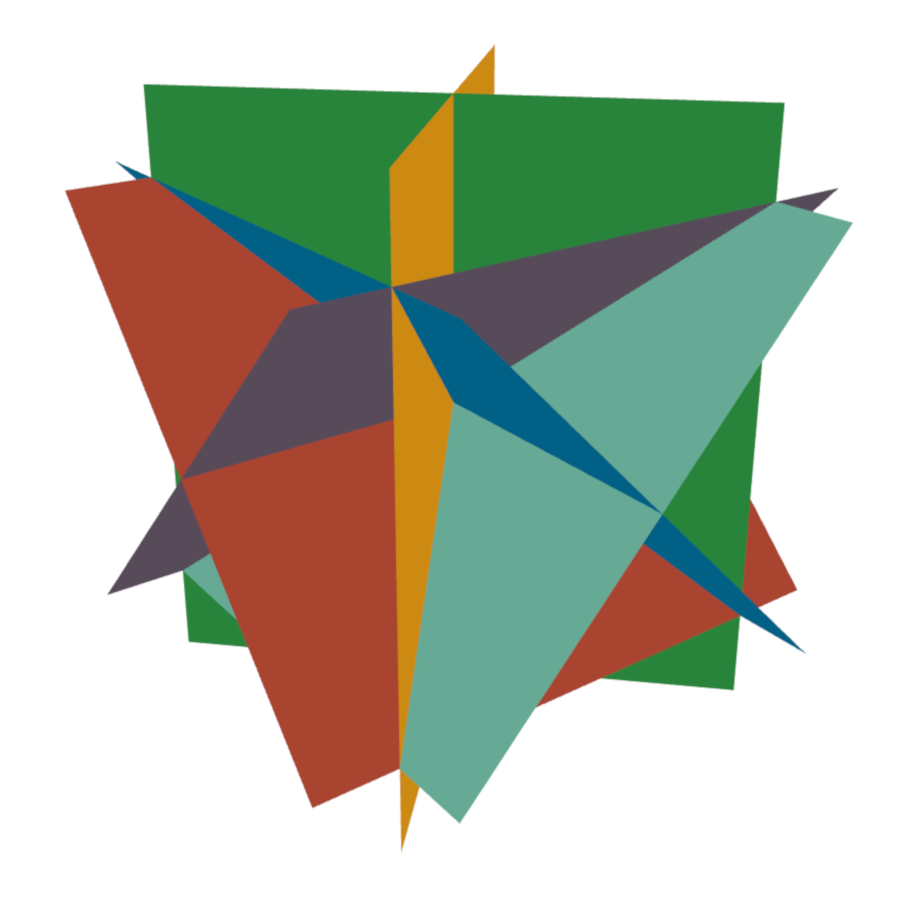}};
            \node (pi4) at (\x + .6, \y + 1.25) {\footnotesize{$\preg{1342}$}};
            \node (pi5) at (\x - .8, \y + 1.3) {\footnotesize{$\preg{1324}$}};
            \node (c2) at (\x + 2.5, \y - 2.3) {};
            \draw[-, color=black] (c1) edge (c2);
            \node[rotate=90] at (.3, \y) {\large{$|C| = 4$}};
        \end{tikzpicture}
    \end{subfigure}%
    \begin{subfigure}[b]{0.32\textwidth}
        \vspace{.9cm}
        \advance\leftskip.8cm
        \includegraphics[height=4.25cm]{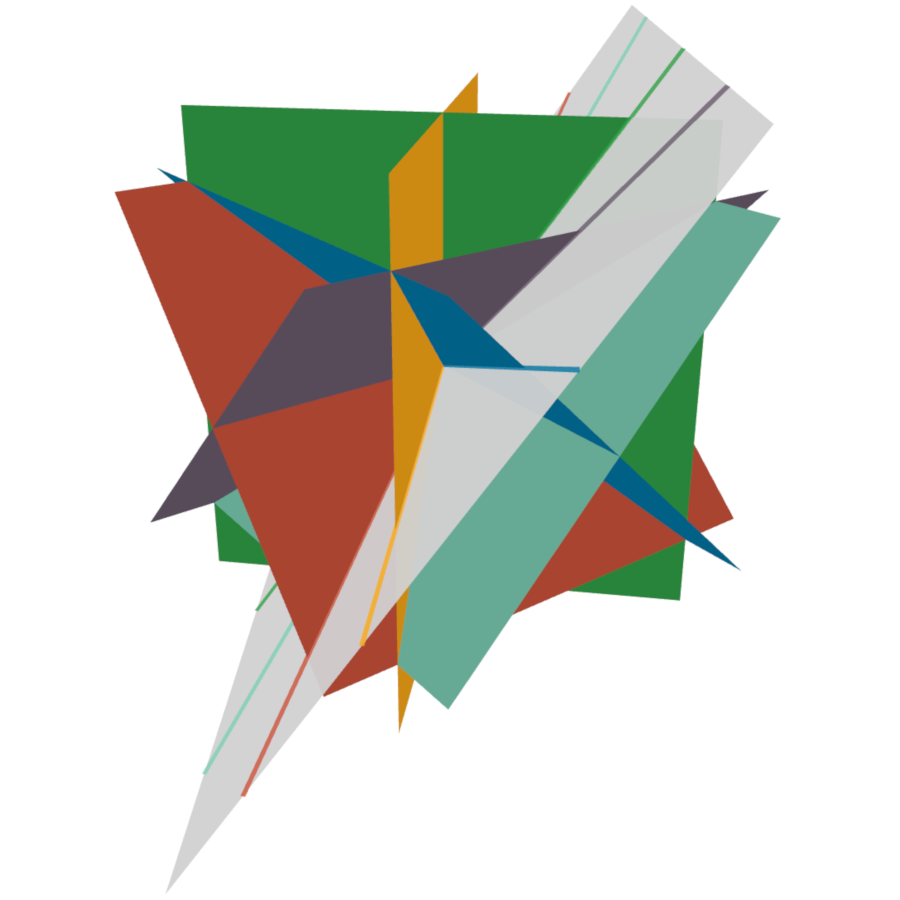}
    \end{subfigure}
    \begin{subfigure}[b]{0.33\textwidth}
       \begin{tikzpicture}[inner sep=0pt, overlay, remember picture] 
            \def\x{2.8}
            \def\y{1.8}
            \node at (\x, \y) {\includegraphics[height=3.7cm]{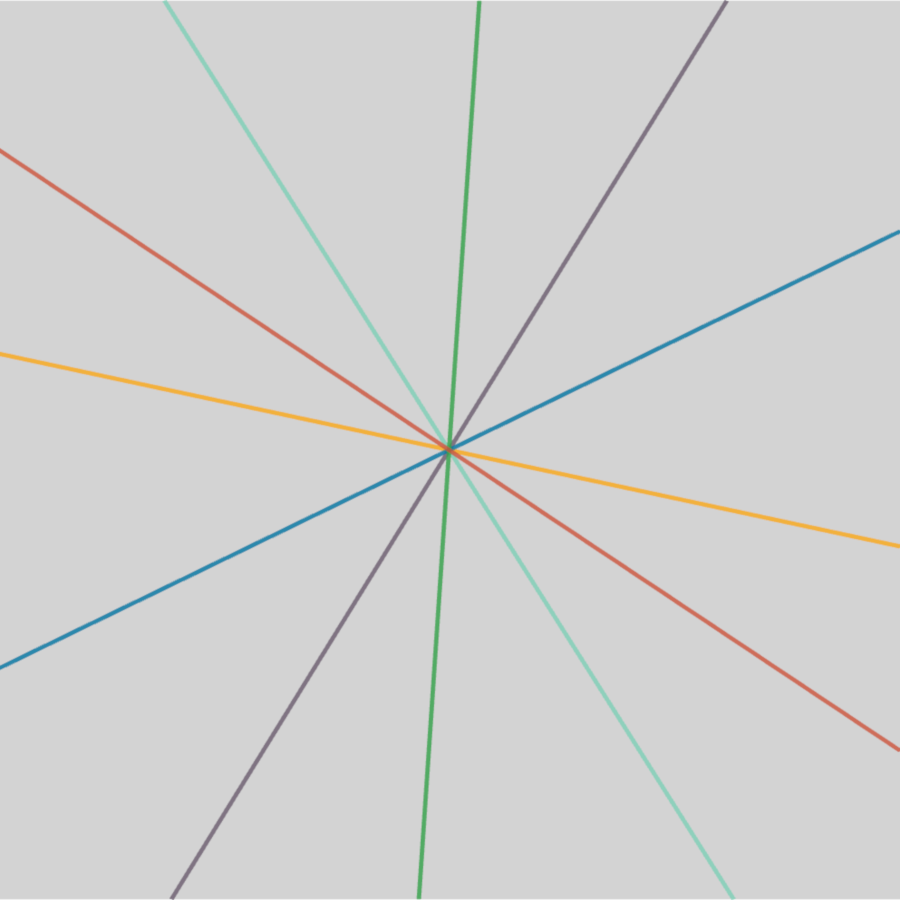}};
            \node (pr) at (3.35, \y + 1.65) {\tiny{$\preg{1342}$}};
            \node (pr2) at (4.1, \y + 1.1) {\tiny{$\preg{1432}$}};
            \node (pr3) at (4.1, \y + .2) {\tiny{$\preg{1423}$}};
            \node (pr4) at (4.1, \y -.5) {\tiny{$\preg{1243}$}};
            \node (pr5) at (4.1, \y - 1.3) {\tiny{$\preg{2143}$}};
            \node (pr6) at (3.25, \y - 1.6) {\tiny{$\preg{2413}$}};
            \node (pr7) at (2.25, \y - 1.65) {\tiny{$\preg{2431}$}};
            \node (pr8) at (1.45, \y - 1.25) {\tiny{$\preg{2341}$}};
            \node (pr9) at (1.45, \y - .15) {\tiny{$\preg{3241}$}};
            \node (pr10) at (1.45, \y + .6) {\tiny{$\preg{3421}$}};
            \node (pr11) at (1.45, \y + 1.4) {\tiny{$\preg{3412}$}};
            \node (pr12) at (2.25, \y + 1.65) {\tiny{$\preg{3142}$}};
            \node (c4) at (-.1 , \y - 1.9) {};
            \node (pl3) at (-1.33, \y + 2.28) {};
            \node (pl4) at (1.85, \y + 1.85) {};
            \draw[->, color=gray] (pl3) edge [bend left=10] (pl4);
            \draw[-, color=black] (c3) edge (c4);
        \end{tikzpicture}
    \end{subfigure}
    \caption{Illustration of Corollary~\ref{cor:stolen} ($3^{rd}$ column) as a result of Observation~\ref{obs:part} ($1^{st}$ column) and Observation~\ref{obs:feas} ($2^{nd}$ column) for $\operatorname{softmax}(\vvec{Wx})$, $\vvec{W} \in \R^{|C| \times d}, \, d=2$. Planes truncated for ease of visualisation. \textbf{Top row}:  In the left column we see the Braid Arrangement for 3 classes partitioning the output space into 6 regions that correspond to permutations: class rankings in increasing order of probability. In the middle column we see that because $d=2$ we can only map $\vvec{x}$ to the feasible logits, a plane (grey) defined by $\vvec{W}$. Therefore, in the right column we see that we can only represent permutations that correspond to the regions we can intersect with this plane. For $|C| = 3$ we can still represent all $6$ rankings of $3$ classes since any plane in general position will intersect all $6$ regions. \textbf{Bottom row}: The Braid Arrangement for $4$ classes. Since $d < |C| - 1$ the plane can only intersect $12$ regions so only $12/24$ permutations are feasible. For example, we see that the plane intersects region $\preg{1342}$ but not $\preg{1324}$ and hence $\perm{1342}$ is feasible while $\perm{1324}$ is not. In fact, the orientation of the plane is such that none of the $6$ $\preg{***4}$ regions are intersected. Therefore $c_4$ cannot be ranked above $c_1, c_2$ and $c_3$ and is unargmaxable as in Figures~\ref{fig:stolen}~and~\ref{fig:stolen2}.}
    \label{fig:braidslice}
\end{figure*}

A class $c_t$ is unargmaxable when all permutations that rank $c_t$ above the rest cannot be realised due to rank constraints.
We explain how this happens by combining the following two observations.

\begin{obs}
\label{obs:part}
\textbf{We can discretise $\R^{|C|}$ into regions corresponding to permutations by segmenting the space with hyperplanes.}
\normalfont

The hyperplanes that partition the output space into regions $\preg{}$ corresponding to permutations are a well known structure in Combinatorics, the \textbf{Braid Hyperplane Arrangement} \citep{stanley2004}.\footnote{See Appendix~\ref{app:braid} for more details on hyperplane arrangements and the Braid Arrangement specifically.} The Braid Arrangement for 3 and 4 classes is illustrated in rows 1 and 2 of Figure~\ref{fig:braidslice} respectively.

In order to be able to rank the classes according to permutation $\preg{}$, our network needs to be able to map
an input $\vvec{x}$ to region $\preg{}$ in the output space.
However, this is not always possible when we have a Softmax Bottleneck as we elaborate below.

\end{obs}

\begin{obs}
\label{obs:feas}
\textbf{When we have rank constraints, only a subspace of $\R^{|C|}$ is feasible}.
\normalfont

\textbf{Case i) $\operatorname{softmax}(\vvec{Wx})$}.
By calculating $\mathbf{y}=\mathbf{Wx}$, the class logits $\vvec{y}$ are a linear combination of $d$ columns of $\vvec{W}$. Therefore, when $d < |C|$ we can only represent a $d$-dimensional subspace of $\R^{|C|}$ at best.
This feasible subspace is illustrated as a grey plane in the middle column of Figure~\ref{fig:braidslice}. 

\textbf{Case ii) $\operatorname{softmax}(\vvec{Wx} + \vvec{b})$}. If we also have a bias term $\vvec{b}$ the model can choose how to offset the subspace. When the bias term $\vvec{b}$ is not in the column space of $\vvec{W}$ the zero vector $\vvec{0}$ is no longer a feasible $\vvec{y}$ and instead of a linear subspace we have an affine subspace. See Figure~\ref{fig:braidslice-bias} in the Appendix for an illustration comparing the two cases.
\end{obs}

\begin{cor}
\label{cor:stolen}
\textbf{A Softmax classifier parametrised by $\vvec{W}$ and $\vvec{b}$ can rank classes in the order of permutation $\bm{\pi}$ iff the affine subspace spanned by $\vvec{W}$ and $\vvec{b}$ intersects region $R_{\bm{\pi}}$ of the Braid Arrangement}.\footnote{This insight of slicing the Braid Arrangement was introduced in~\citet{kamiya2011}.} 
\normalfont
When \textbf{$d < |C| - 1$} there are regions that cannot be intersected.\footnote{When $d = C -1$ we can still intersect all regions, because the Braid Arrangement always has rank $|C| - 1$ (all its normal vectors are perpendicular to the all ones vector $\vvec{1}$). 
}
The feasible permutations in our example correspond to the regions formed on the grey plane illustrated in the rightmost column of Figure~\ref{fig:braidslice}. Note that for $|C| = 4$ only 12 out of 24 regions can be intersected.
\end{cor}

As we make the Softmax Bottleneck narrower by reducing the dimension $d$ of the Softmax inputs, more permutations become infeasible~\citep{good1977, kamiya2005}. Importantly, if we choose $|C|$ and $d$ and whether to use a bias term, changing the values of the Softmax weights changes the set of feasible permutations but not the cardinality of the set~\citep{Cover1967,Smith2014}. See Appendix~\ref{app:numregions} for more details. 

\begin{cor}
\textbf{Class $c_t$ is unargmaxable when any permutation that would rank class $c_t$ above all other classes is infeasible.}
\end{cor}

\subsubsection{Effect of Softmax Bias Term}

\label{sec:softbias}

Without a bias term the regions corresponding to permutations are unbounded (see the rightmost column of Figure~\ref{fig:braidslice}). As such, imposing any range restrictions on the Softmax layer inputs $\vvec{x}$ does not change the feasible regions as long as the restriction includes the origin.
However, when we introduce a bias term we also get bounded regions (see Figure~\ref{fig:braidslice-bias} in the Appendix that contrasts the two situations). Therefore, in this case the scale of the inputs to the Softmax layer also matters. If the inputs do not have a large enough range, there will be regions that exist but cannot be reached by the feature encoder.

\subsection{Exact Algorithm}
\label{sec:exact}
Given a softmax layer parametrised by $\vvec{W}$ and $\vvec{b}$, are there any classes that are unargmaxable? We first describe a slow, but exact algorithm to answer this question.

An exact algorithm will either prove class $c_t$ is argmaxable by returning a feasible point $\vvec{x}: \argmax{(\vvec{W}\vvec{x} + \vvec{b})} = c_t$ or it will prove $c_t$ is unargmaxable by verifying no such point exists.

To check if a region exists that ranks $c_t$ above all others, we need to find an input $\vvec{x} \in \R^d$ that satisfies the following constraints:
\begin{equation}
\label{eq:braid}
\resizebox{.99 \columnwidth}{!}{%
$P(c_{i} \mid \vvec{x}) < P(c_{t} \mid \vvec{x}), \quad \forall i: \, 1 \leq i \leq |C|,\, i \neq t$%
}
\end{equation}
Each of the above constraints is equivalent to restricting $\vvec{x}$ to a halfspace (see Appendix \ref{app:halfspace}).
Hence, if all above inequalities are enforced, $\vvec{x}$ is restricted to an intersection of halfspaces.

\begin{equation}
\label{eq:braidplanes}
\begin{aligned}
(\vvec{w}_{c_i} - \vvec{w}_{c_t}) \T \vvec{x} + (b_{c_i} - b_{c_t}) < 0 \\
\forall i:\quad  1 \leq i \leq |C|, \quad i \neq t
\end{aligned}
\end{equation}
If the intersection of halfspaces is empty, there is no $\vvec{x}$ for which class $c_t$ can be ranked above all others - and hence $c_t$ is unargmaxable. We can find a point in an intersection of halfspaces via linear programming, albeit we found this algorithm to be slow in practice for $|C| > 1000$.

\subsubsection{Chebyshev Center Linear Programme}

The Chebyshev center of a polytope~\citep[p. 417]{boyd2004} is the center of the largest ball of radius $r$ that can be embedded within the polytope. We can find the Chebyshev center $\vvec{x}$ and the radius $r$  with the following linear programme.
\begin{alignat}{2}
& \text{maximise} & \quad &  r  \nonumber \\
& \text{subject to} &  & \vvec{w}_{i}\T \vvec{x} + r \Vert \vvec{w}_i \Vert_2 \leq -b_i, \scriptstyle \quad  1 \leq i \leq |C| - 1 \nonumber \\
&                   &  & \vvec{x} \leq 100 \nonumber \\
&                   &  & \vvec{x} \geq -100 \nonumber \\
&                   &  & r > 0
\end{alignat}
Where $\vvec{w}_i = \vvec{w}_{c_i} - \vvec{w}_{c_t}$ and $b_i = b_{c_i} - b_{c_t}, \, \forall i : c_i \neq c_t$.
We further constrain $\vvec{x}$ to guarantee the regions are bounded, since the Chebyshev center is not defined otherwise. This constraint also captures the fact that neural network activations are not arbitrarily large.

If the above linear programme is feasible, we know that class $c_t$ is argmaxable and we also get a lower bound on the volume of the region for which it is solvable by inspecting $r$. On the other hand, if the linear programme is infeasible, $c_t$ is unargmaxable.

\subsection{Approximate Algorithm}
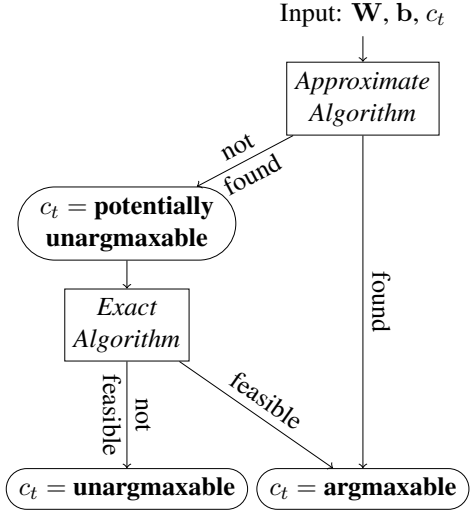
\begin{figure}
  \centering
  \scalebox{0.9}{
      \begin{tikzpicture}
      \def \xs {1.5}
      \def \ys {2.3}
      \node[](a) at (1*\xs, -1.3) {Input: $\vvec{W}$, $\vvec{b}$, $c_t$};
      \node[draw, align=center](approx) at (1*\xs, -1.1*\ys) {\textit{Approximate}\\\textit{Algorithm}};
      \node[draw, align=center](exact) at (-1.3*\xs, -2.55*\ys) {\textit{Exact}\\\textit{Algorithm}};
      \node[draw, rounded rectangle, align=center](approxb) at (-1.3*\xs, -1.9*\ys) {$c_t =$ \textbf{potentially} \\ \textbf{unargmaxable}};
      \node[draw, rounded rectangle](no) at (1*\xs, -3.6*\ys) {$c_t =$ \textbf{argmaxable}};
      \node[draw, rounded rectangle](yes) at (-1.3*\xs, -3.6*\ys) {$c_t =$ \textbf{unargmaxable}};
      
      \draw [->] (a) edge (approx) ;
      
      \draw [->] (approx) to node [above, sloped] {found} (no) ;
      \draw [->, align=center] (approx) to node [sloped, pos=0.5] {not\\found} (approxb) ;
      \draw [->, align=center] (approxb) to node [sloped, pos=0.5] {} (exact) ;
      
      \draw [->] (exact) to node [above, sloped] {feasible} (no) ;
      \draw [->, align=center] (exact) to node [sloped, pos=0.5] {not\\feasible} (yes) ;
      \end{tikzpicture}
  }
  \caption{Algorithm to verify whether class $c_t$ is argmaxable. We first run the approximate algorithm, which quickly proves most vocabulary tokens are argmaxable. If it fails to find a solution in $N$ steps, we rely on the exact algorithm to either find a solution or prove there is no solution, meaning $c_t$ is unargmaxable.}
  \label{fig:pipeline}
\end{figure}

The exact algorithm was too slow to run for the whole vocabulary. 
In order to avoid running the exact algorithm for every single vocabulary item, we developed an incomplete algorithm~\citep{kautz2009} with a one-sided error, which can quickly rule out most tokens, leaving only a small number to be checked by the exact algorithm. It proves that $c_t$ is \textbf{argmaxable} by finding an input $\vvec{x}$ for which $c_t$ has the largest activation.
Unlike the exact algorithm, if no solution exists it cannot prove that the token is \textbf{unargmaxable}. 
Hence, we terminate our search after a predetermined number of steps.
We denote any tokens not shown to be argmaxable by the approximate algorithm as \textbf{potentially unargmaxable} and we run the exact algorithm on them.
An illustration of the way we combine the exact and approximate algorithms to decide whether class $c_t$ is argmaxable can be seen in Figure~\ref{fig:pipeline}.

\subsubsection{Braid Reflect}
The idea behind this approximate algorithm is to use the Braid Hyperplane Arrangement as a map to guide us towards a point $\vvec{x}$ for which $c_t$ has the largest activation.
\begin{figure}[t]
\removelatexerror
\begin{algorithm}[H]
    \SetAlgoLined
    \DontPrintSemicolon
    \KwData{Class index $c_t$, $\vvec{x} \in \R^{d}$, $\vvec{W} \in \R^{|C| \times d},\, \vvec{b} \in \R^{|C|}$}
    $c_i$ = argmax($\vvec{W}\vvec{x} + \vvec{b}$)\;
    $\vvec{w} = (\vvec{w}_{c_t} - \vvec{w}_{c_i}) \T$\;\label{alg:braid}
    $b = b_{c_t} - b_{c_i}$\;\label{alg:braidb}
    $\vvec{w'} = \frac{\vvec{w}}{\norm{\vvec{w}}_{2}}$\; \label{alg:dist1}
    $d = \vvec{w'}\T \vvec{x}$\;\label{alg:dist2}
    $\vvec{x} = \vvec{x} - 2 (d + \frac{b}{\norm{\vvec{w}}_2}) \vvec{w'}$\;\label{alg:reflect}
 \caption{Braid reflection step}
\end{algorithm}
\caption{Move $\vvec{x}$ to region where $P(c_t) > P(c_i)$.}
\label{fig:braidswap}
\end{figure}
To show that class $c_t$ is argmaxable, it suffices to find an input $\vvec{x}$ for which the largest probability is assigned to $c_t$. Empirically we found this to be easy for most classes. 

We begin by interpreting the actual weight vector as the candidate input $\vvec{x} = \vvec{w}_{c_t}\T$. We do so since the dot product of two vectors is larger when the two vectors point in the same direction.\footnote{$\vvec{a}\T\vvec{b} = \norm{\vvec{a}}_2 \norm{\vvec{b}}_2 \cos{\theta}\,$ is maximised for $\theta = 0$} While the magnitude of the vectors affects the dot product, we found the above initialisation worked well empirically. When $c_t$ is not the argmax for $\vvec{x}$ and $c_i$ is instead, Relation~\ref{eq:braidplanes} for $c_i$ and $c_t$ will have the wrong sign. The sign of this relation defines which side of the Braid hyperplane for $c_i$ and $c_t$ we are on. To correct the sign, we construct the normal vector and offset of the Braid hyperplane (Lines \ref{alg:braid}, \ref{alg:braidb} in Figure~\ref{fig:braidswap}), compute the distance of $\vvec{x}$ from it (Line \ref{alg:dist2}), and reflect $\vvec{x}$ across it (Line \ref{alg:reflect}).\footnote{When no offset is involved, the reflection operation is the Householder transformation~\citep{householder1958}.} We repeat the above operation until either $c_t$ is the argmax or we have used up our budget of $N$ steps.

\section{Experiments}

In this Section we use the combined algorithm from Figure~\ref{fig:pipeline} to search models for unargmaxable tokens.

We test $7$ LMs and $143$ MT models. We find that unargmaxable tokens only occur in $13$ MT models, but these are mostly infrequent and noisy vocabulary tokens. We therefore do not expect such tokens to affect translation quality per se.

We also find that nearly all vocabulary tokens of LMs and student MT models can be verified with less than $N=10$ steps of the approximate algorithm. In contrast, other MT models need thousands of steps and also rely on the exact algorithm. In this sense, models that need fewer steps are easier to verify: the search problem for their arrangement of Softmax weights is easier.

Throughout the following experiments we assumed the Softmax inputs were bounded in magnitude for all dimensions $-100 \leq x_i \leq 100$.
As we mentioned in Subsection~\ref{sec:softbias}, if we have a Softmax bias term, there are bounded regions.
If the bounded regions are large, even though the outputs are not theoretically bounded, they are practically bounded since neural network feature encoders cannot produce arbitrarily large activations and some regions may be unreachable\footnote{The validity of our assumption is only relevant for models we find to be bounded. We therefore verified that $-100 \leq \vvec{x} \leq 100$ holds for two of them, see Appendix~\ref{app:range}.}.
For the approximate algorithm, we search for a solution with a patience of $N=2500$ steps and resort to the exact algorithm if the approximate method fails or returns a point outside the aforementioned bounds.
We use Gurobi~\citep{gurobi} as the linear programme solver.
We accessed the model parameters either via NumPy~\cite{harris2020} or PyTorch~\citep{torch2019}.
The experiments took 3 days to run on an AMD 3900X $12$-core CPU using $10$ threads and 64Gb of RAM.

\subsection{Language Models (0/7 Unargmaxable)}

We checked $7$ widely used LMs for unargmaxable tokens. While some of these models such as BERT~\citep{devlin2018} are not directly used for generation, a recent trend is to use these large LMs as prompt models~\citep{Liu2021} for few shot learning. A prompt model obviates the need for a separate classifier by rephrasing a classification task as slot filling given a task specific template. Prompt approaches commonly choose the answer for the slot by argmaxing the Softmax distribution obtained by a LM. Hence we verify that there are no answers that are unargmaxable.

BERT, RoBERTa~\citep{Liu2019}, XLM-RoBERTa~\citep{conneau-etal-2020-roberta} and GPT2~\citep{Radford2019} did not exhibit any unargmaxable tokens and can be assessed without resorting to the exact algorithm (see Table~\ref{tab:bounded-repr} in the Appendix). Moreover, the LMs were very easy to verify with the approximate algorithm requiring less than $1.2$ steps per token on average.

\subsection{Machine Translation (13/143 Unargmaxable)}

\begin{table}[h]
\centering
    \scalebox{0.65}{
    \begin{tabular}{l c c c c}
    \toprule
    model source & Helsinki & FAIR & Edinburgh & Bergamot \\
    \midrule
    unargmaxable & 13/32 & 0/4 & 0/82 & 0/25 \\
    dataset & OPUS & WMT'19 & WMT'17 & multiple\footnotemark \\
    architecture & Transf & Transf & LSTM & Transf \\
    feature dim $d$ & 512 & 1024 & 500,512 & 256,512,1024 \\
    Softmax bias & \cmark & \xmark & \cmark & \cmark \\
    tied embeds & enc+dec+out & dec+out & dec+out & enc+dec+out\\
    \bottomrule
    \end{tabular}
    }
    \caption{Results for the MT models we verified.}
    \label{tab:stats}
\end{table}
\footnotetext{\href{https://github.com/browsermt/students}{https://github.com/browsermt/students}}

In the case of MT models, the feature encoder comprises the whole encoder-decoder network excluding the last layer of the decoder.
We first focus on models which we found to have unargmaxable tokens and then briefly describe models that did not.
A summary of the results and characteristics of the models we checked can be seen in Table~\ref{tab:stats}.
More detailed results can be found in Tables ~\ref{tab:bounded-helsinki-opus}, ~\ref{tab:bounded-FAIR}, ~\ref{tab:bounded-bergamot} and ~\ref{tab:bounded-ediwmt} in the Appendix.

\textbf{Helsinki NLP OPUS (13/32 Unargmaxable).}
The $32$ models we use for this subset of experiments are MT models released through Hugging Face~\citep{wolf-etal-2020-transformers}.
We use models introduced in~\citet{opusmt}. These models are trained on subsets of OPUS. All models are transformer models trained using Marian~\citep{junczys-dowmunt2018}. They include a bias term,  have a tied encoder and decoder and $d=512$.

Unargmaxable tokens, if present, will affect generation in the target language. We therefore restrict our analysis to the target language vocabulary. To facilitate this, we inspect translation models for which the source and target languages have different scripts. We explore $32$ models with source and target pairs amongst Arabic (ar), Hebrew (he), English (en), German (de), French(fr), Spanish (es), Finnish (fi), Polish (pl), Greek (el), Russian (ru), Bulgarian (bg), Korean (ko) and Japanese (ja).
We rely on the script to disambiguate between source and target language and discard irrelevant tokens from other languages. We also ignore vocabulary tokens containing digits and punctuation.

\begin{figure}[t]
    \includegraphics[width=\columnwidth]{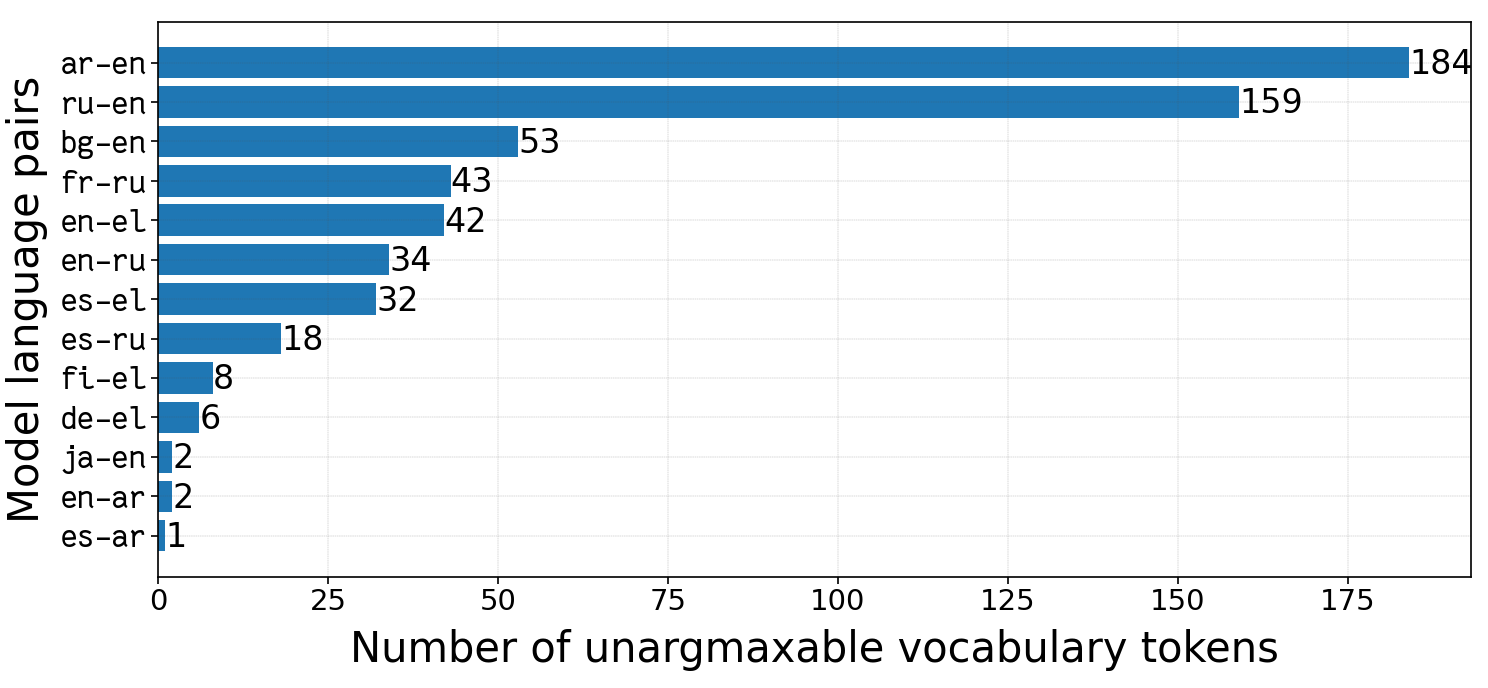}
    \caption{$13/32$ HelsinkiNLP models have vocabulary tokens that cannot be predicted using greedy decoding.}
    \label{fig:helsinki}
\end{figure}

In Figure~\ref{fig:helsinki} we can see the number of Byte Pair Encoding \citep[BPE;][]{sennrich2016} tokens that were unargmaxable for these models, sorted in decreasing order.
As can be seen, all tokens are argmaxable for $19/32$ language pairs. For the remaining $13$ languages, while there can be quite a few unargmaxable tokens, most would not be expected to affect translation quality.

Out of the set of $427$ unique unargmaxable BPE tokens, $307/476$ are single character subword tokens and only $2$ are word stem BPE segments: \textit{erecti} (bg-en) and \textit{\foreignlanguage{russian}{Предварительны}} (en-ru) which means ``preliminary'' in Russian. The rest include the \textit{<unk>} token and noisy subword unicode tokens such as \foreignlanguage{russian}{ќЌЌќ}, \foreignlanguage{greek}{ὶῖῖ} and \foreignlanguage{greek}{ἀὐῇ}.

On closer inspection of the SentencePiece tokeniser we found that both \textit{\foreignlanguage{russian}{Предварительны}} and \textit{erecti} come up as tokenisation alternatives that make them rare and irregular.
We found that the \textit{\foreignlanguage{russian}{Предварительны}} token was rare since it is capitalised and only occurs once, while another occurrence was caused by a BPE segmentation corner case due to Unicode token variation of \textit{\foreignlanguage{russian}{Предварительны-e}}. Other mentions having \textit{\foreignlanguage{russian}{Предварительны}} as a substring were split differently. In a similar vein, we found that the \textit{erecti} token occurred due to BPE corner cases for \textit{erecti-0-n}, \textit{erecti-lis-)}, \textit{erecti-l}, \textit{erecti-.} and \text{erecti-cle} many of which are misspellings or rare word forms from clinical text. As such, the impact of these tokens being unargmaxable is small since there are alternative ones the MT model can prefer over them which could even correct spelling mistakes.

\textbf{FAIR WMT'19 (0/4 Unargmaxable).}
We checked 4 FAIR models (en-ru, ru-en, en-de, de-en) submitted to WMT'19~\citep{ng-etal-2019-facebook}.
These transformer models have $d=1024$ and do not employ a Softmax bias term.

None of the FAIR models were found to have unargmaxable tokens, but for some tokens we had to rely on the exact algorithm to show this.

\textbf{Edinburgh WMT'17 (0/82 Unargmaxable).}
These WMT'17 submissions~\citep{sennrich-etal-2017-university} were ensembles of left-to-right trained models (l2r) and right-to-left trained models (r2l).
These were LSTMs trained with Nematus using $d=500$ or $d=512$ and Softmax weights tied with the decoder input embeddings.
The models include a bias term.

None of the models have unargmaxable tokens.
However, we found that models that comprise an ensemble varied a lot in how easy it was to show that the vocabulary was argmaxable, despite them differing solely in the random seed used for weight initialisation.
As an example, zh-en.l2r(1) had $8$ tokens that needed to be verified with the exact algorithm, zh-en.l2r(2) had  $3$ and zh-en.l2r(3) had $366$.
This highlights that random initialisation alone is enough to lead to very different arrangements of Softmax weights.

\textbf{Bergamot (0/25 Unargmaxable).}
The Bergamot project\footnote{\href{https://browser.mt}{https://browser.mt}} model repository contains both large transformer-base and transformer-big teacher models, as well as small knowledge distilled~\citep{kim-rush-2016-sequence} student models.
Student models have $d=256$ (tiny) or $d=512$ (base), while teacher models have $d=1024$.
Interestingly, we find that it is easier to show that student models are argmaxable when compared to teacher models, despite student models having Softmax weights $1/2$ or $1/4$ the dimensions of the teacher model.

\section{Discussion}
We conclude from our experiments that  \textit{it is possible to have unargmaxable tokens, but this rarely occurs in practice} for tokens that would lead to irrecoverable errors in the MT models we checked. A limitation of our conclusions is that beam search is usually preferred over greedy decoding for MT models used in practice. We leave the question of whether unargmaxable tokens also impact beam search for future work.

It is challenging to make exact claims about what can cause tokens to be unargmaxable because the models we tested varied in so many ways.
However, we outline some general trends below.

\subsection{Infrequent Tokens Are the Victims}
The most general observation is that the tokens that are more likely to be unargmaxable or are hard to prove to be argmaxable are the infrequent ones.
This can be seen in Figures~\ref{fig:ensemble-diff} and~\ref{fig:teacher-student} in the Appendix, where the x-axis contains the vocabulary of the models sorted left to right by increasing frequency. Each dot represents the number of steps needed to check whether a token is argmaxable or not, and as can be seen the values to the right are generally much higher than those to the left.

This result is in line with previous work that highlights the limitations of the Softmax layer when modelling rare words for LM~\citep{chen2016,labeau2019} and MT~\citep{nguyen2017,raunak-2020} and infrequent classes for image classification~\citep{Kang2020}.

\subsection{Some Models Are Easier to Verify}
We found that the LMs and student MT model vocabularies can be shown to be argmaxable with one step of the approximate algorithm on average. On the other hand, for Helsinki NLP and FAIR MT models more than $10$ steps were needed.

To put the above observations into context, we also check the behaviour of our algorithms on randomly initialised parameters.
If we initialise a Softmax layer of $|C|=10000$ classes using a uniform distribution $U(-1, 1)$ we do not expect unargmaxable tokens to exist after $d=30$ (see Figure~\ref{fig:random-uniform} in the Appendix). Moreover, any randomly initialised parameters can be checked using the approximate algorithm with fewer steps as we increase $d$.

From this perspective, it is surprising that student models were easier to show to be argmaxable than the teacher models, despite the Softmax weight dimensionality of the student models being much lower (256 for tiny, versus 1024 for teacher).
This shows that effective neural MT models do not need to be hard to check, but nevertheless neural models trained on the original data can sometimes converge to such an arrangement of weights.

\section{Conclusions and Future Work}

In this work we discretised the outputs of Softmax and showed how dimensionality constraints shrink the set of feasible class rankings and can lead to some classes being impossible to predict using argmax.
In our experiments we demonstrated that while MT models can have unargmaxable vocabulary tokens, this does not occur often in our experiments. Moreover, for the models we tested the unargmaxable tokens would not create discernible differences in translation quality as the tokens are noisy and infrequent.
We release an algorithm to detect whether some classes are unargmaxable with the hope that this will be helpful to the wider community working on a plethora of different models where the observed phenomena may vary.

In future work, we aim to investigate any learnability consequences more closely. As we saw, when using an approximate search algorithm, it is much harder to find argmaxable classes in some models than it is in others. Since gradient descent algorithms are also iterative search algorithms seeking optimal parameters, we hypothesise that it will be challenging to train neural network encoders to map activations to regions of the input space that a search algorithm cannot find easily. 
Hence, although some tokens may not be provably unargmaxable because of constraints imposed by the Softmax parameters of the last layer, some tokens may still be very hard to produce because of difficulties encountered by the feature encoder. To this end, a more holistic investigation into the consequences of the loss in expressivity in low-rank classifiers is warranted.

\section*{Broader Impact}

Unargmaxability directly impacts fairness, since certain model outputs, further from being underrepresented, may not be represented at all. As we discussed, low-rank classifiers have limited expressivity compared to full rank classifiers, and thus have to explicitly choose which rankings of classes to retain feasible when using argmax prediction.
As such, by choosing to use a low-rank model, we allow the data and training procedure to specify which rankings should remain feasible, and harmful biases in our data can be propagated and further exacerbated~\citep{hooker2021} by our models due to unargmaxability.
For example, it could be the case that underrepresented groups find no representation in the outputs of such models when such relevant outputs are unargmaxable. As researchers, we should be aware of this limitation when choosing how to
parametrise our models~\citep{hooker2019} and seek to either control such phenomena or verify models are not harmful before moving them from research to production.

In addition to the above considerations, linear classification layers are vulnerable to targeted attacks via data poisoning techniques~\citep{goldblum2020}, especially under the scenario where shared models are used as feature extractors~\citep{ji2018}.
A subset of such techniques, known as feature collisions~\citep{shafahi2018,goldblum2020}, exploit the arrangement of the training examples in feature space to force the misclassification of a target example.
Attacks such as Convex Polytope~\citep{zhu2019a} and Bullseye Polytope~\citep{aghakhani2021}, specifically target the unargmaxability weakness~\citep{Cover1967,demeter2020} we elaborated on in the paper.
While such attacks assume they are able to inject examples into a training set used for fine-tuning, this is not an unrealistic assumption.
This is especially true for recommender systems, where adversarial attacks can create fake users such that a target item is removed from a target user's top-k list~\citep{christakopoulou2019}.

\section*{Acknowledgements}
We thank Seraphina Goldfarb-Tarrant, Elizabeth Nielsen and Sabine Weber for help with languages, Beatrice Alex, Sameer Bansal, Panagiotis Eustratiadis, Sharon Goldwater, Chantriolnt-Andreas Kapourani, Oli Liu, Sue Liu, Yevgen Matusevych, Kate McCurdy, Boris Mitrovic, Mark-Jan Nederhof, Laura Perez-Beltrachini, Evi Potsi, Marina Potsi, Jesse Sigal, Mark Steedman, Ivan Titov and Sabine Weber for feedback and support, Antonio Vergari for feedback, guidance and tirelessly discussing low-rank constraints and Shay Cohen for insightful suggestions and pointers to OEIS. We also thank David Demeter for an extensive discussion on Stolen Probability and the anonymous reviewers for helpful questions and comments.

\lettrine[image=true, lines=2, findent=1ex, nindent=0ex, loversize=.15]{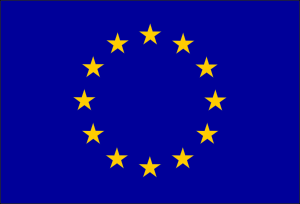}%
This work was supported by the Engineering and Physical Sciences Research Council [grant number EP/R513209/1] and 
\ProjectType\ \textit{\ProjectName}, which has received funding from the European Union's Horizon 2020 research and innovation programme under grant agreement No \GrantNo.

\bibliography{main}
\bibliographystyle{acl_natbib}
\appendix

\begin{figure*}
    \begin{subfigure}[b]{0.24\textwidth}
        \includegraphics[height=3.8cm]{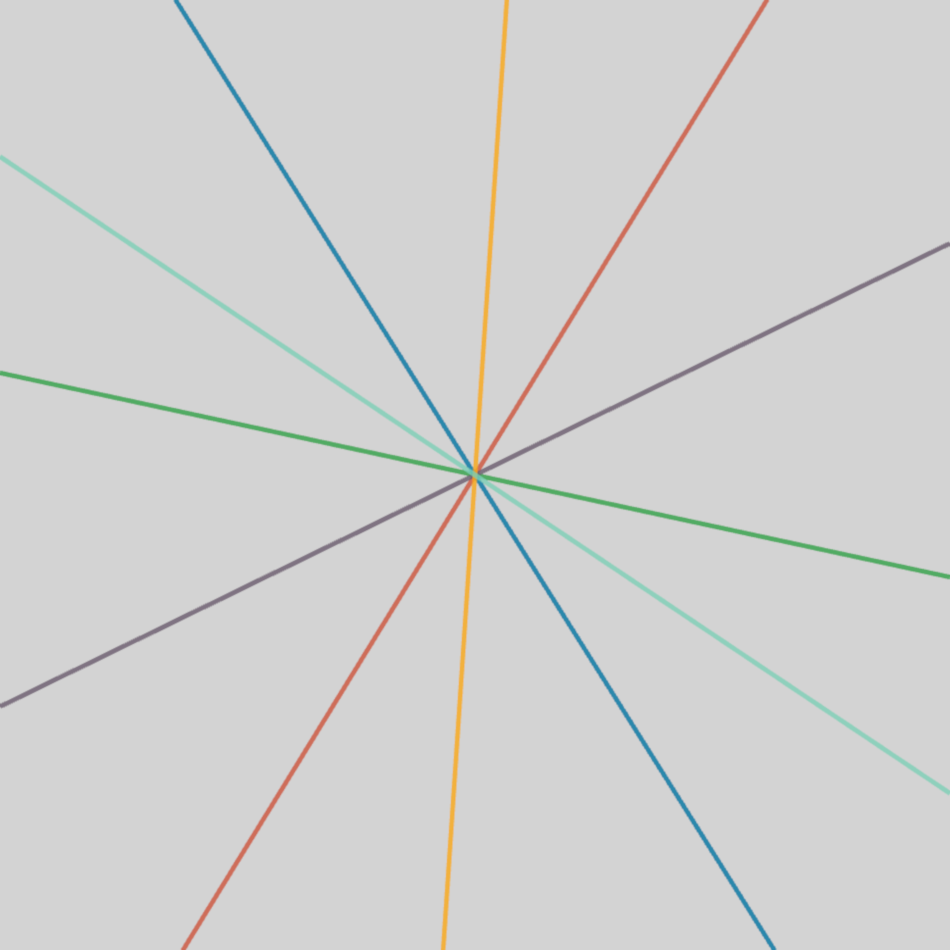}
        \caption{Input space $\vvec{b}=\vvec{0}$}
    \end{subfigure}%
    \begin{subfigure}[b]{0.24\textwidth}
        \includegraphics[height=4cm]{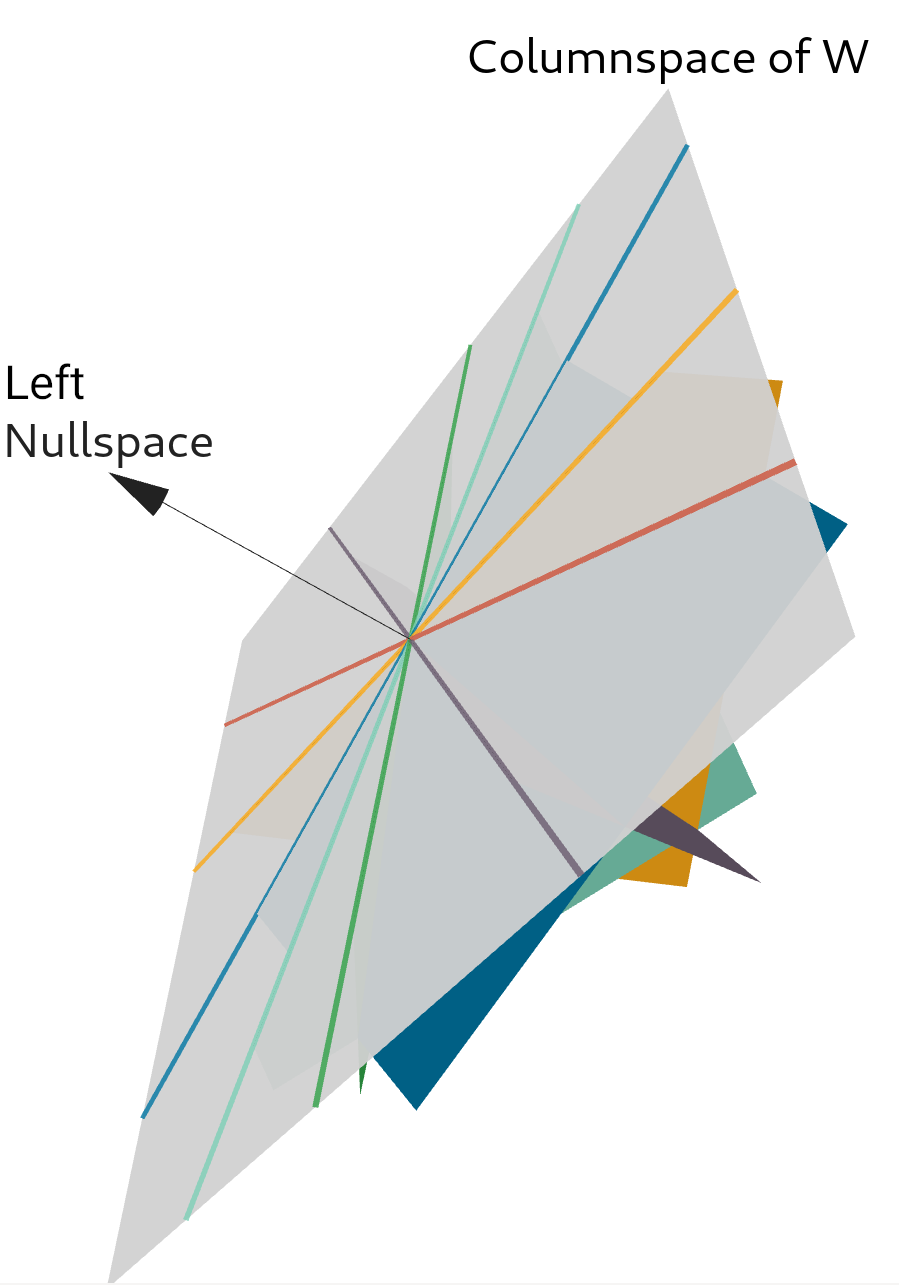}
        \caption{Output space $\vvec{b}= \vvec{0}$}
    \end{subfigure}
    \hspace{.03\textwidth}
    \begin{subfigure}[b]{0.24\textwidth}
        \includegraphics[height=3.8cm]{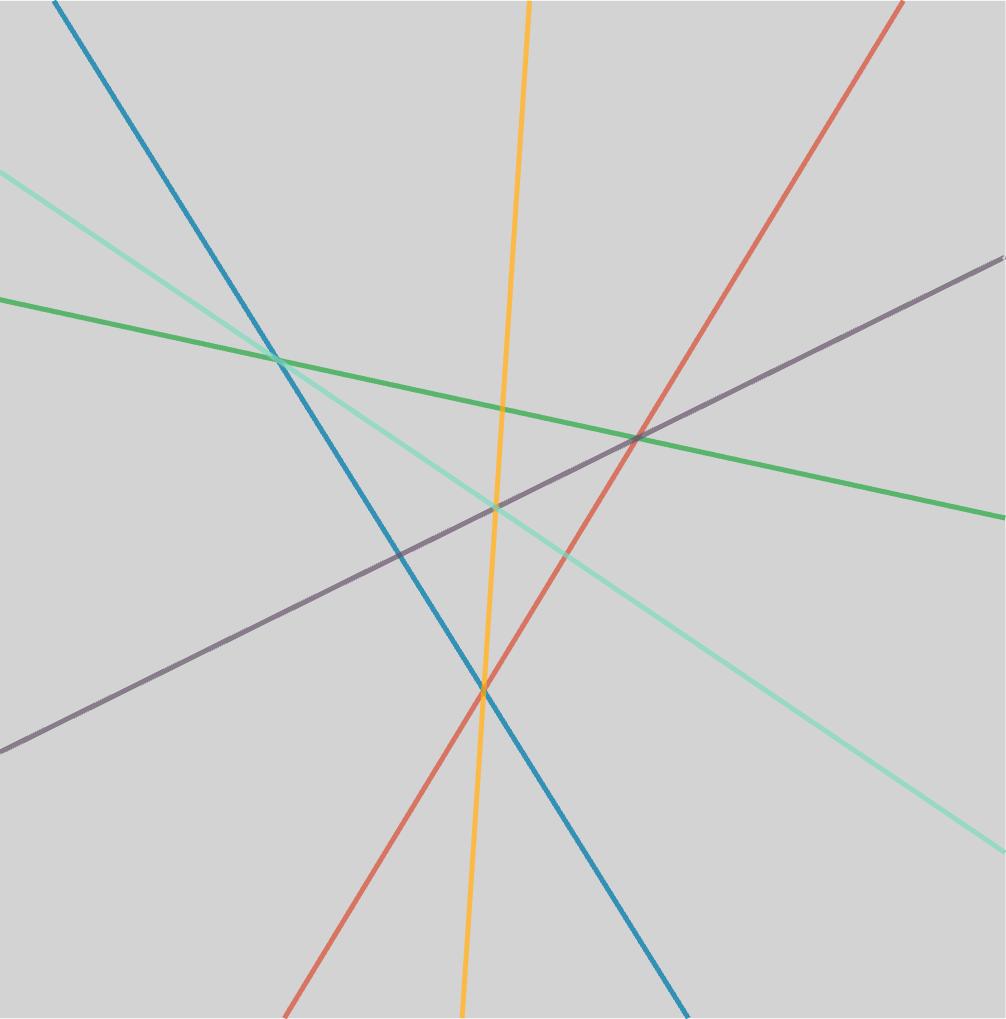}
        \caption{Input space $\vvec{b}\neq \vvec{0}$}
    \end{subfigure}%
    \begin{subfigure}[b]{0.24\textwidth}
        \includegraphics[height=4cm]{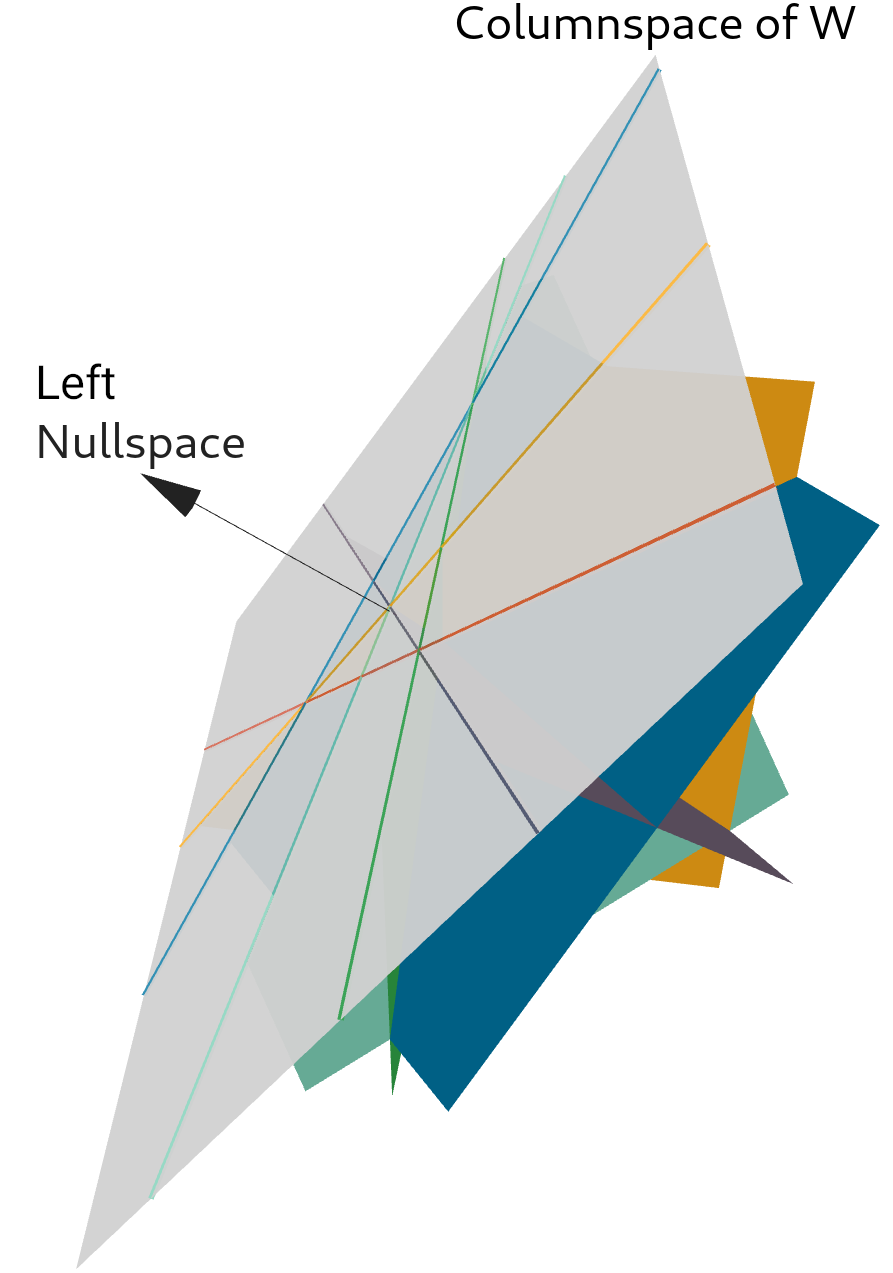}
        \caption{Output space $\vvec{b}\neq \vvec{0}$}
    \end{subfigure}
    \caption{Effect of bias term $\vvec{b}$ on feasible permutations of $\operatorname{softmax}(\vvec{Wx} + \vvec{b})$, $\vvec{W} \in \R^{|C| \times d}, \, d=2, \, |C|=4$.   Having a bias term offsets the grey plane and allows it to not pass through the origin. This increases the number of regions by creating bounded regions seen in Subfigures c and d. 
    Each region intersected by the grey 2D plane corresponds to a feasible permutation. We therefore obtain $18/24$ feasible permutations if we include a bias term, compared to $12/24$ without one.}
    \label{fig:braidslice-bias}
\end{figure*}

\section{Halfspace interpretation}
\label{app:halfspace}

As promised, here is the derivation showing that if $P(c_i \mid \vvec{x}) < P(c_j \mid \vvec{x})$ then $\vvec{x}$ is constrained to a halfspace.

We have:
\begin{equation}
\begin{aligned}
P(c_i \mid \vvec{x}) < P(c_j \mid \vvec{x}) & \iff \\
\frac{e^{\vvec{w}_{c_i} \T \vvec{x} + b_{c_i}}}{\sum_{i'} e^{\vvec{w}_{c_{i'}} \T \vvec{x} + b_{c_{i'}}}} <
\frac{e^{\vvec{w}_{c_j} \T \vvec{x} + b_{c_j}}}{\sum_{i'} e^{\vvec{w}_{c_{i'}} \T \vvec{x} + b_{c_{i'}}}} & \iff \\
e^{\vvec{w}_{c_i} \T \vvec{x} + b_{c_i}} < e^{\vvec{w}_{c_j} \T \vvec{x} + b_{c_j}} & \iff \\
\frac{e^{\vvec{w}_{c_i} \T \vvec{x} + b_{c_i}}}{e^{\vvec{w}_{c_j} \T \vvec{x} + b_{c_j}}} < 1 & \iff \\
e^{(\vvec{w}_{c_i} - \vvec{w}_{c_j}) \T \vvec{x} + (b_{c_i} - b_{c_j})} < e^0 & \iff \\
(\vvec{w}_{c_i} - \vvec{w}_{c_j}) \T \vvec{x} + (b_{c_i} - b_{c_j}) < 0 & \\
\end{aligned}
\end{equation}
$\vvec{x}$ is therefore constrained to a halfspace defined by normal vector $\vvec{w}_{c_i} - \vvec{w}_{c_j}$ and offset by $b_{c_i} - b_{c_j}$.
This linear form defined by the normal vector and offset is the ``shadow'' in the input dimension of our friend, the Braid Arrangement, as we will make clear in the next Section (see Derivation~\ref{eq:proj}).

\section{Hyperplane Arrangements}
\label{app:braid}

Excellent resources to learn more about hyperplane arrangements are \citet{stanley2004} and \href{http://math.sfsu.edu/federico/Clase/Polytopes/polytopes.html}{Federico Ardila's lectures on polytopes (see Lecture 34 onwards)}. Connections between hyperplane arrangement theory and Machine Learning can be found in \citet[Chapter~40]{mackay2004}. For those who prefer a more gentle introduction via a hands on approach, Sagemath~\cite{sagemath} contains implementations of many hyperplane arrangements and functions that we found useful when learning this material. We give a brief introduction to hyperplane arrangements below.

A \emph{hyperplane} in a vector space $\R^d$ is an affine subspace of dimension $d-1$. The hyperplane $\mathcal{H}$ has one degree of freedom removed by specifying a constraint: a normal vector $\mathbf{w} \in \R^d$ to which it is perpendicular. The hyperplane may also be offset by $b$ in that direction $\mathcal{H} = \{\mathbf{x} \in \R^d : \mathbf{w}\T\mathbf{x} = b\}$.

A \emph{real hyperplane arrangement} $\mathcal{A}$ is defined as a set of $n$ hyperplanes in $\R^d$, $\mathcal{A} = \{\mathcal{H}_1, \mathcal{H}_2 \ldots \mathcal{H}_n \}$. The set of \emph{regions} $\mathcal{R}$ defined by a hyperplane arrangement $\mathcal{A}$ are the connected components $X$ of Euclidean space $\R^d$ left when we remove the hyperplanes $\mathcal{A}$, namely $X = \R^d - \bigcup_{\mathcal{H} \in \mathcal{A}} \mathcal{H}$.  As an example, Subfigure (a) in Figure \ref{fig:braidslice-bias} has 12 regions while Subfigure (c) has 18 regions.

\subsection{Braid Arrangement}
The Braid Arrangement $\mathcal{B}_n$ is a hyperplane arrangement that partitions space into $n!$ regions corresponding to permutations.
It can be constructed in $\R^{n}$ from the standard basis, the rows of the identity matrix $\vvec{I}$, $\vvec{e}_i = \row_i(\vvec{I})\T $, $\mathbf{e}_i \in \R^n$, by taking all $n \choose 2$ pairs of differences between them, each difference defining the normal vector of a hyperplane $\mathcal{H}_{i, j}$ of the Braid Arrangement.
\begin{equation}
\begin{aligned}
\mathcal{B}_n = \{\mathcal{H}_{i, j} \quad \forall i, j : \, 1 \leq i < j \leq n\}, \\
\mathcal{H}_{i, j} = \{\mathbf{x} \in \R^n : (\mathbf{e}_i - \mathbf{e}_j)\T\mathbf{x} = 0\}
\end{aligned}
\end{equation}
The Braid Arrangement for $n=3$ and $n=4$ can be seen in Figure~\ref{fig:braidslice}. It has $n \choose 2$ hyperplanes, one per pair of dimensions in $\R^n$. Hence there are $3$ hyperplanes for $|C| = 3$ and $6$ hyperplanes for $|C| = 4$. As an example, when we have $4$ classes the normal vector for $\mathcal{H}_{1, 3}$ is $\vvec{w}_{1,3} =\begin{bmatrix} 1 & 0 & -1 & 0 \end{bmatrix}\T$.
As can be verified by taking the dot product $\vvec{w}_{i, j}\T \vvec{x}$, the result is positive if $x_i$ > $x_j$ and negative if vice versa.
Therefore, each hyperplane bisects space into two regions one for each possible ranking of the pair of coordinates.

To see how the hyperplanes intersect to give us a region $\preg{}$, we express a permutation (total order) over $|C|$ classes, such as that in Relation~\ref{eq:ranking}, using a chain of $|C| - 1$ pairwise inequalities.
\begin{equation}
P(c_{\pi_i} \mid \vvec{x}) < P(c_{\pi_{i+1}} \mid \vvec{x}), \quad 1 \leq i \leq |C| - 1
\end{equation}
Each above constraint is equivalent to choosing a side of a braid hyperplane. By imposing all constraints, we obtain a region $\preg{}$ as the intersection of $|C| - 1$ halfspaces.
There is therefore bijection between permutations and regions of the Braid Arrangement $\perm{} \leftrightarrow \preg{}$.

\subsection{Restricting the Braid Arrangement to Lower Dimensions}

In the Softmax layer of a neural network we often compute the output space activations $\vvec{y} \in \R^n$ by applying a final affine layer to the Softmax input space $\vvec{x} \in \R^d$.
\begin{equation}
\vvec{y} = \vvec{Wx} + \vvec{b},\quad \vvec{W} \in \R^{n \times d},\, \vvec{b} \in \R^{n}
\end{equation}
What do the Braid Arrangement hyperplanes look like in the input dimension $d$? Let us start from the output space $\R^n$ and work backwards towards the input space $\R^d$.
\begin{align}
y_i < y_j \implies & (\vvec{e}_i - \vvec{e}_j)\T \vvec{y} < 0  \nonumber \\
            & \vvec{e}_i\T \vvec{y} - \vvec{e}_j\T \vvec{y} < 0   \nonumber \\
            & \vvec{e}_i\T (\vvec{Wx} + \vvec{b}) - \vvec{e}_j\T (\vvec{Wx} + \vvec{b}) < 0   \nonumber \\
            & \vvec{w}_i\T \vvec{x} + b_i - \vvec{w}_j\T \vvec{x} - b_j < 0   \nonumber \\
            & (\vvec{w}_i - \vvec{w}_j) \T \vvec{x} + (b_i - b_j) < 0
\label{eq:proj}
\end{align}
We therefore see that if $d < n$ we can think of how the Braid Arrangement classifies outputs into permutations from two equivalent perspectives:
\begin{itemize}
    \item In the output space $\R^n$ not all $\vvec{y}$ are feasible, we can only classify an input $\vvec{x}$ as a permutation $\perm{}$ if the affine layer can map $\vvec{x}$ to $\preg{}$. This can be seen in Subfigures b and d of Figure~\ref{fig:braidslice-bias} where the feasible outputs are a plane that intersects the Braid Arrangement.
    \item In the input space $\R^d$ all $\vvec{x}$ are feasible but we only see the projection of the Braid Arrangement in this lower dimension. This can be seen in Subfigures a and c of Figure~\ref{fig:braidslice-bias}.
\end{itemize}

\begin{figure}[h!]
\includegraphics[width=\columnwidth]{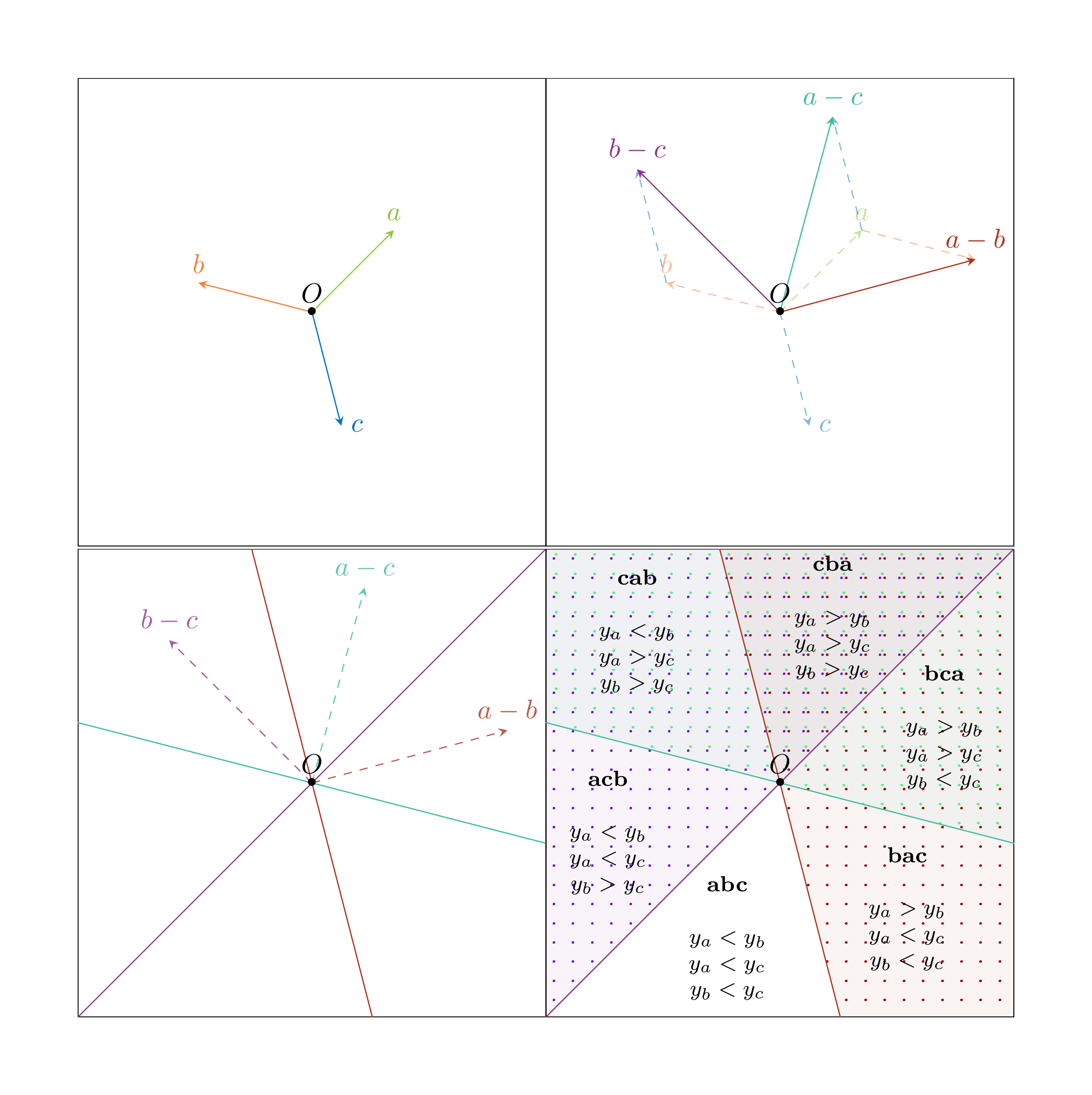}
\caption{Constructing the Braid Arrangement in the input space for $|C| = 3$ classes and $d=2$. \emph{Top left}: The Softmax weights $\vvec{W} \in \R^{|C| \times d}$ for $3$ classes, $a,b,c$. Each vector is a row of the weight matrix. \emph{Top right}: We form the normal vectors for the braid hyperplanes by taking all pairs of differences between the basis vectors. \emph{Bottom left}: The Braid hyperplanes are perpendicular to the normal vectors. Each hyperplane bisects space into two regions, one comprises the set of $\vvec{x}$ for which class $i$ has a larger activation that class $j$ and the second vice versa. \emph{Bottom right}: The hyperplanes partition space into $3!=6$ regions corresponding to permutations. Each permutation contains the indices that sort the activations over classes in increasing order. Softmax decision boundaries are unions of two regions, e.g. regions \textbf{cba} and \textbf{bca} for class \textbf{a}.}
\label{fig:braiddiff}
\end{figure}

The construction of the Braid Arrangement in the input space is illustrated in Figure~\ref{fig:braiddiff}, albeit without the bias term.

\section{Number of Regions (Feasible Permutations) of the Restricted Braid Arrangement}
\label{app:numregions}
The number of feasible permutations is invariant to specific choices of $\vvec{W}$ and $\vvec{b}$ \citep{Cover1967,Smith2014} and only depends on the dimensionality of the softmax inputs $d$, the number of classes $|C|$ and whether we specify a bias term $\vvec{b}$ not in the columnspace of $\vvec{W}$. Namely, the cardinality of the set of feasible permutations does not change, but the members of the set do - they depend on the specific values in $\vvec{W}$ and $\vvec{b}$. There exists a recurrence formula to obtain the number of feasible permutations for a particular $|C|$ and $d$~\cite{good1977, kamiya2005}. See our code and the relations in ~\citep{Smith2014} for more details.

\subsection{Softmax with no Bias Term}
The number of feasible permutations as a function of $|C|$ and $d$ when we have a Softmax with no bias term can be seen in Table~\ref{tab:numregions-nobias}.
When $d \geq |C| - 1$ all permutations corresponding to ways of ranking $|C|$ classes are feasible (table cells with $d = |C| - 1$ are highlighted in bold). However, as we make the Softmax Bottleneck narrower, we can represent less permutations, as can be seen from the numbers reported below the diagonal.

        \begin{table*}
        \scalebox{0.9}{
        \begin{tabular}{cl| llllllllll }
        \toprule
        \multicolumn{2}{c}{} 
                & \multicolumn{ 10 }{c}{\textsc{Bottleneck dimensionality $d$}}  \\
        \multicolumn{2}{c}{} 
         & 1 & 2 & 3 & 4 & 5 & 6 & 7 & 8 & 9 & 10 \\
        \midrule
        \multirow{ 9 }{*}{\rotatebox[origin=c]{90}{\textsc{Number classes $|C|$}}}

                & 2 & \hh{2} & 2 & 2 & 2 & 2 & 2 & 2 & 2 & 2 & 2\\
                & 3 & 2 & \hh{6} & 6 & 6 & 6 & 6 & 6 & 6 & 6 & 6\\
                & 4 & 2 & \textcolor{orange}{\textit{12}} & \hh{24} & 24 & 24 & 24 & 24 & 24 & 24 & 24\\
                & 5 & 2 & 20 & 72 & \hh{120} & 120 & 120 & 120 & 120 & 120 & 120\\
                & 6 & 2 & 30 & 172 & 480 & \hh{720} & 720 & 720 & 720 & 720 & 720\\
                & 7 & 2 & 42 & 352 & 1512 & 3600 & \hh{5040} & 5040 & 5040 & 5040 & 5040\\
                & 8 & 2 & 56 & 646 & 3976 & 14184 & 30240 & \hh{40320} & 40320 & 40320 & 40320\\
                & 9 & 2 & 72 & 1094 & 9144 & 45992 & 143712 & 282240 & \hh{362880} & 362880 & 362880\\
                & 10 & 2 & 90 & 1742 & 18990 & 128288 & 557640 & 1575648 & 2903040 & \hh{3628800} & 3628800\\

        \bottomrule
        \end{tabular}
        }
        \caption{Number of permutation regions defined by a bottlenecked Softmax layer $Softmax(\vvec{W}x)$ with no bias term. When $d \geq |C| - 1$ all permutations corresponding to ways of ranking $|C|$ classes are feasible. $12$ in italics corresponds to the number of regions shown in the left Subfigure of Figure~\ref{fig:braidslice-bias}. \href{https://oeis.org/A071223}{https://oeis.org/A071223}.}
        \label{tab:numregions-nobias}
        \end{table*}

\subsection{Softmax with Bias Term}

The number of feasible permutations as a function of $|C|$ and $d$ when we have a Softmax with a bias term is larger as can be seen in Table~\ref{tab:numregions-bias}.
As we saw in Figure~\ref{fig:braidslice-bias}, this is because a bias term can offset the representible linear subspace to an affine subspace which can intersect more regions of the Braid Arrangement.

        \begin{table*}
        \scalebox{0.9}{
        \begin{tabular}{cl| llllllllll }
        \toprule
        \multicolumn{2}{c}{} 
                & \multicolumn{ 10 }{c}{\textsc{Bottleneck dimensionality $d$}}  \\
        \multicolumn{2}{c}{} 
         & 1 & 2 & 3 & 4 & 5 & 6 & 7 & 8 & 9 & 10 \\
        \midrule
        \multirow{ 9 }{*}{\rotatebox[origin=c]{90}{\textsc{Number classes $|C|$}}}
                & 2 & \hh{2} & 2 & 2 & 2 & 2 & 2 & 2 & 2 & 2 & 2\\
                & 3 & 4 & \hh{6} & 6 & 6 & 6 & 6 & 6 & 6 & 6 & 6\\
                & 4 & 7 & \textcolor{orange}{\textit{18}} & \hh{24} & 24 & 24 & 24 & 24 & 24 & 24 & 24\\
                & 5 & 11 & 46 & 96 & \hh{120} & 120 & 120 & 120 & 120 & 120 & 120\\
                & 6 & 16 & 101 & 326 & 600 & \hh{720} & 720 & 720 & 720 & 720 & 720\\
                & 7 & 22 & 197 & 932 & 2556 & 4320 & \hh{5040} & 5040 & 5040 & 5040 & 5040\\
                & 8 & 29 & 351 & 2311 & 9080 & 22212 & 35280 & \hh{40320} & 40320 & 40320 & 40320\\
                & 9 & 37 & 583 & 5119 & 27568 & 94852 & 212976 & 322560 & \hh{362880} & 362880 & 362880\\
                & 10 & 46 & 916 & 10366 & 73639 & 342964 & 1066644 & 2239344 & 3265920 & \hh{3628800} & 3628800\\

        \bottomrule
        \end{tabular}
        }
        \caption{Number of permutation regions defined by a bottlenecked Softmax layer $Softmax(\vvec{W}x + \vvec{b})$. When $d \geq |C| - 1$ all permutations corresponding to ways of ranking $|C|$ classes are feasible. $18$ in italics corresponds to the number of regions shown in the right Subfigure of Figure~\ref{fig:braidslice-bias}.}
        \label{tab:numregions-bias}
        \end{table*}
        
\FloatBarrier

\section{Braid Reflect Approximate Algorithm}
\begin{figure}[h]
\removelatexerror
\begin{algorithm}[H]
    \SetAlgoLined
    \DontPrintSemicolon
    \KwData{Class index $c_t$, $\vvec{W} \in \R^{|C| \times d},\, \vvec{b} \in \R^{|C|}$}
    \KwResult{Whether $c_t$ is unargmaxable}
        unargmaxable = true\;
        patience = 2500\;
        $\vvec{x} = \vvec{w_{c_t}\T}$\;
        \While{patience}{
            $c_i$ = argmax($\vvec{W}\vvec{x} + \vvec{b}$)\;
            \eIf{$c_i = c_t$}{
                unargmaxable = false\;
                \Break
                }{
                $\vvec{w} = (\vvec{w}_{c_t} - \vvec{w}_{c_i}) \T$\;
                $b = b_{c_t} - b_{c_i}$\;
                $\vvec{w'} = \frac{\vvec{w}}{\norm{\vvec{w}}_{2}}$\;
                $d = \vvec{w'}\T \vvec{x}$\;
                $\vvec{x} = \vvec{x} - 2 (d + \frac{b}{\norm{\vvec{w}}_2}) \vvec{w'}$\;
                patience = patience - 1\;
            }
        }
 \caption{Braid reflect}
\end{algorithm}
\caption{Approximate algorithm to detect whether class $c_t$ is unargmaxable.}
\label{fig:braidswapfull}
\end{figure}

\FloatBarrier
\section{Unargmaxable Token Search Results}

\begin{table}[h!]
\scalebox{0.85}{
\begin{tabular}{c | c c}
\toprule
\thead{model} & \thead{\# potentially\\ unargmaxable} & \thead{\# unargmaxable}\\
\midrule
          bert-base-cased & 0 & 0 \\
          bert-base-uncased & 0 & 0 \\
          roberta-base & 0 & 0 \\
          roberta-large & 0 & 0 \\
          xlm-roberta-base & 0 & 0 \\
          xlm-roberta-large & 0 & 0 \\
          gpt2 & 0 & 0 \\
\bottomrule
\end{tabular}
}
\caption{Unargmaxable token search results for LMs. \textbf{potentially unargmaxable} is the number of tokens that the approximate algorithm failed to prove were argmaxable. \textbf{argmaxable} is the number of unargmaxable tokens according to the exact algorithm. No tokens were found to be unargmaxable.}
\label{tab:bounded-repr}
\end{table}

\begin{table}[h!]
\scalebox{0.75}{
\begin{tabular}{c c | c c}
\toprule
\thead{source} & \thead{model} & \thead{\# potentially\\ unargmaxable} & \thead{\# unargmaxable}\\
\midrule

\multirow{32}{*}{\shortstack{Helsinki\\NLP}}
         &  opus-mt-ja-en & 109 & 2 \\
         &  opus-mt-ru-en & 90 & 159 \\
         &  opus-mt-bg-en & 93 & 53 \\
         &  opus-mt-ja-en(2) & 14 & 0 \\
         &  opus-mt-ar-en & 40 & 184 \\
         &  opus-mt-en-el & 75 & 42 \\
         &  opus-mt-de-el & 115 & 6 \\
         &  opus-mt-ar-el & 41 & 0 \\
         &  opus-mt-es-el & 67 & 32 \\
         &  opus-mt-fi-el & 57 & 8 \\
         &  opus-mt-ar-he & 3 & 0 \\
         &  opus-mt-de-he & 4 & 0 \\
         &  opus-mt-es-he & 3 & 0 \\
         &  opus-mt-fr-he & 1 & 0 \\
         &  opus-mt-fi-he & 7 & 0 \\
         &  opus-mt-ja-he & 0 & 0 \\
         &  opus-mt-en-ar & 21 & 2 \\
         &  opus-mt-el-ar & 12 & 0 \\
         &  opus-mt-es-ar & 17 & 1 \\
         &  opus-mt-fr-ar & 17 & 0 \\
         &  opus-mt-he-ar & 7 & 0 \\
         &  opus-mt-it-ar & 8 & 0 \\
         &  opus-mt-ja-ar & 4 & 0 \\
         &  opus-mt-pl-ar & 52 & 0 \\
         &  opus-mt-ru-ar & 8 & 0 \\
         &  opus-mt-en-ru & 98 & 34 \\
         &  opus-mt-es-ru & 42 & 18 \\
         &  opus-mt-fi-ru & 1 & 0 \\
         &  opus-mt-fr-ru & 34 & 43 \\
         &  opus-mt-he-ru & 5 & 0 \\
         &  opus-mt-ja-ru & 13 & 0 \\
         &  opus-mt-ko-ru & 2 & 0 \\
\bottomrule
\end{tabular}
}
\caption{Unargmaxable token search results for Helsinki NLP OPUS models. \textbf{potentially unargmaxable} is the number of tokens that the approximate algorithm failed to prove were argmaxable. \textbf{unargmaxable} is the number of unargmaxable tokens according to the exact algorithm. For 13/32 models some infrequent tokens were found to be unargmaxable.}
\label{tab:bounded-helsinki-opus}
\end{table}

\begin{table}[h!]
\scalebox{0.75}{
\begin{tabular}{c c | c c}
\toprule
\thead{source} & \thead{model} & \thead{\# potentially\\ unargmaxable} & \thead{\# unargmaxable}\\
\midrule
\multirow{4}{*}{\shortstack{FAIR}}
         &  facebook/wmt19-en-ru & 5 & 0 \\
         &  facebook/wmt19-ru-en & 64 & 0 \\
         &  facebook/wmt19-de-en & 173 & 0 \\
         &  facebook/wmt19-en-de & 184 & 0 \\

\bottomrule
\end{tabular}
}
\caption{Unargmaxable token search results for FAIR WMT'19 models. \textbf{potentially unargmaxable} is the number of tokens that the approximate algorithm failed to prove were argmaxable. \textbf{unargmaxable} is the number of unargmaxable tokens according to the exact algorithm. No tokens were found to be unargmaxable.}
\label{tab:bounded-FAIR}
\end{table}

\begin{table}[h!]
\scalebox{0.7}{
\begin{tabular}{c l | c c}
\toprule
\thead{source} & \thead{model} & \thead{\# potentially\\ unargmaxable} & \thead{\# unargmaxable}\\
\midrule
\multirow{25}{*}{\shortstack{Bergamot}}
         &  cs-en.student.base & 0 & 0 \\
         &  es-en.teacher.bigx2(1) & 0 & 0 \\
         &  es-en.teacher.bigx2(2) & 0 & 0 \\
         &  en-es.teacher.bigx2(1) & 0 & 0 \\
         &  en-es.teacher.bigx2(2) & 0 & 0 \\
         &  et-en.teacher.bigx2(1) & 2 & 0 \\
         &  et-en.teacher.bigx2(2) & 1 & 0 \\
         &  en-et.teacher.bigx2(1) & 1 & 0 \\
         &  en-et.teacher.bigx2(2) & 1 & 0 \\
         &  nb-en.teacher.base & 0 & 0 \\
         &  nn-en.teacher.base & 0 & 0 \\
         &  is-en.teacher.base & 0 & 0 \\
         &  cs-en.student.base & 0 & 0 \\
         &  cs-en.student.tiny11 & 0 & 0 \\
         &  en-cs.student.base & 0 & 0 \\
         &  en-cs.student.tiny11 & 0 & 0 \\
         &  en-de.student.base & 0 & 0 \\
         &  en-de.student.tiny11 & 0 & 0 \\
         &  es-en.student.tiny11 & 0 & 0 \\
         &  en-es.student.tiny11 & 0 & 0 \\
         &  et-en.student.tiny11 & 0 & 0 \\
         &  en-et.student.tiny11 & 0 & 0 \\
         &  is-en.student.tiny11 & 0 & 0 \\
         &  nb-en.student.tiny11 & 0 & 0 \\
         &  nn-en.student.tiny11 & 0 & 0 \\
\bottomrule
\end{tabular}
}
\caption{Unargmaxable token search results for Bergamot models. \textbf{potentially unargmaxable} is the number of tokens that the approximate algorithm failed to prove were argmaxable. \textbf{unargmaxable} is the number of unargmaxable tokens according to the exact algorithm. No tokens were found to be unargmaxable. Interestingly, student models were much easier to prove argmaxable than teacher models, despite student model Softmax weights being lower dimensional.}
\label{tab:bounded-bergamot}
\end{table}

\begin{table}[h!]
\scalebox{0.75}{
\begin{tabular}{c c | c c}
\toprule
\thead{source} & \thead{model} & \thead{\# potentially\\ unargmaxable} & \thead{\# unargmaxable}\\
\midrule
\multirow{30}{*}{\shortstack{WMT'17\\Edinburgh}}
         &  en-cs.l2r(1-4) & $\leq$ 2 & 0 \\
         &  en-cs.r2l(1-4) & $\leq$ 1 & 0 \\
         &  cs-en.l2r(1-4) & $\leq$ 2 & 0 \\
         &  cs-en.r2l(1-4) & 0 & 0 \\
         &  en-de.l2r(1-4) & $\leq$ 1 & 0 \\
         &  en-de.r2l(1-4) & $\leq$ 2 & 0 \\
         &  de-en.l2r(1-4) & $\leq$ 2 & 0 \\
         &  de-en.r2l(1-4) & 0 & 0 \\
         &  en-ru.l2r(1-4) & 0 & 0 \\
         &  ru-en.l2r(1-4) & 0 & 0 \\
         &  ru-en.r2l(1-4) & 0 & 0 \\
         &  en-tr.l2r(1-4) & $\leq$ 5 & 0 \\
         &  en-tr.r2l(1-4) & $\leq$ 4 & 0 \\
         &  lv-en.l2r(1-4) & 0 & 0 \\
         &  lv-en.r2l(1-4) & $\leq$ 1 & 0 \\
         &  tr-en.l2r(1) & 2 & 0 \\
         &  tr-en.l2r(2) & 8 & 0 \\
         &  tr-en.l2r(3) & 6 & 0 \\
         &  tr-en.l2r(4) & 2 & 0 \\
         &  tr-en.r2l(1) & 4 & 0 \\
         &  tr-en.r2l(2) & 0 & 0 \\
         &  tr-en.r2l(3) & 6 & 0 \\
         &  tr-en.r2l(4) & 4 & 0 \\
         &  en-zh.l2r(1) & 3 & 0 \\
         &  en-zh.l2r(2) & 3 & 0 \\
         &  en-zh.l2r(3) & 14 & 0 \\
         &  en-zh.l2r(4) & 1 & 0 \\
         &  en-zh.r2l(1) & 2 & 0 \\
         &  en-zh.r2l(2) & 0 & 0 \\
         &  en-zh.r2l(3) & 7 & 0 \\
         &  en-zh.r2l(4) & 7 & 0 \\
         &  zh-en.l2r(1) & 8 & 0 \\
         &  zh-en.l2r(2) & 3 & 0 \\
         &  zh-en.l2r(3) & 366 & 0 \\
         &  zh-en.r2l(1-3) & $\leq$ 3 & 0 \\
\bottomrule
\end{tabular}
}
\caption{Unargmaxable token search results for Edinburgh WMT'17 submission (ensemble) models. \textbf{potentially unargmaxable} is the number of tokens that the approximate algorithm failed to prove were argmaxable. \textbf{unargmaxable} is the number of unargmaxable tokens according to the exact algorithm. \textbf{r2l} and \textbf{l2r} refer to training direction, with \textbf{l2r} denoting training left to right and \textbf{r2l} right to left.  Models submitted were ensembles, hence there are more than one model per language pair and direction. When all models per language pair and direction had less than $5$ counts, we summarise all models with a single row, e.g. (1-4).}
\label{tab:bounded-ediwmt}
\end{table}

\FloatBarrier

\begin{figure*}[h!]
\includegraphics[width=\textwidth]{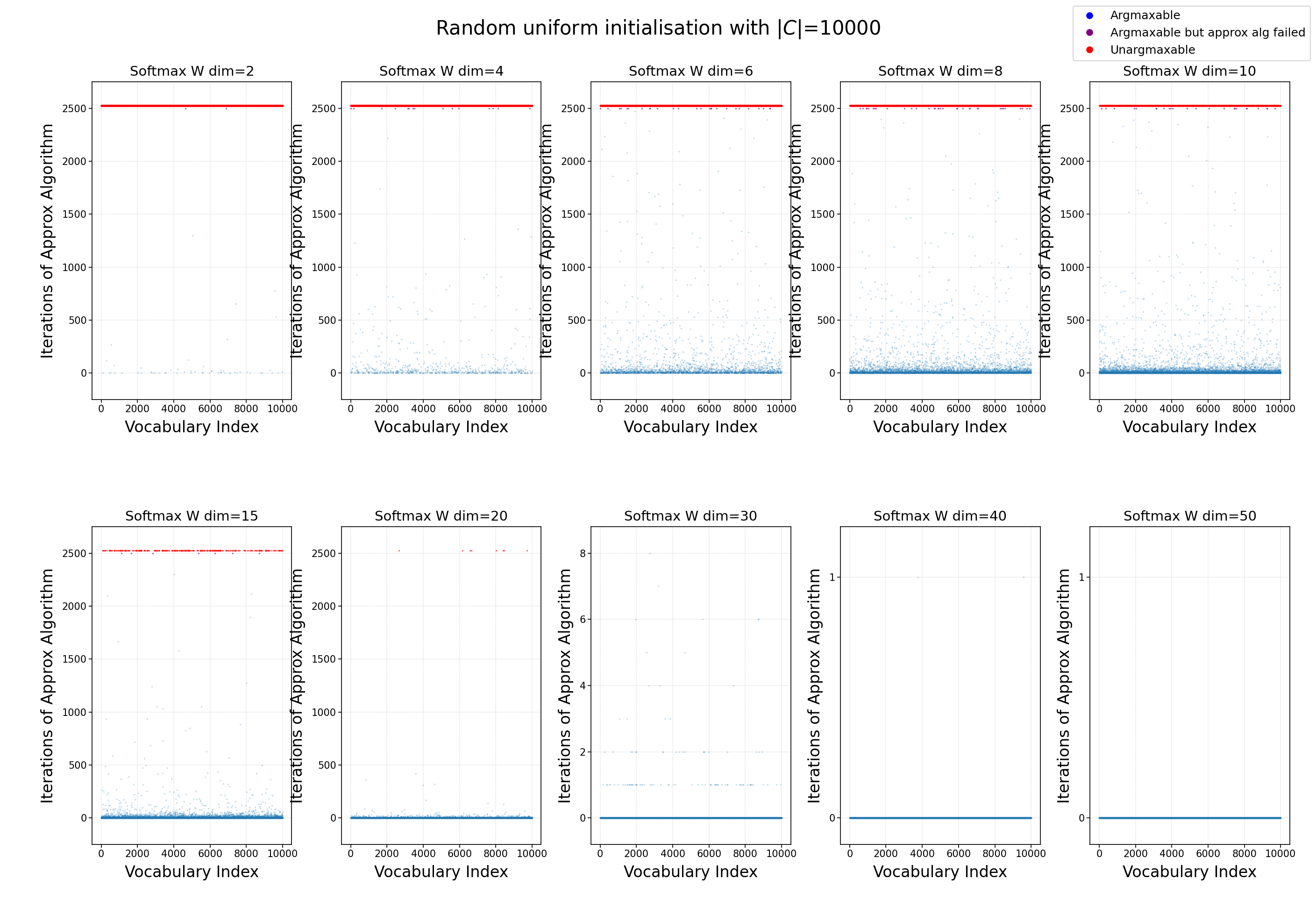}
\caption{Illustration of Softmax weight dimensionality affecting the number of unargmaxable tokens when weights are randomly initialised for a vocabulary of 10000. The Softmax weights and bias term are initialised using a uniform $U(-1, 1)$ distribution. Unargmaxable tokens are unlikely to occur as we increase the dimensionality of the weight vectors. This can be seen in the subplots from top-left to bottom-right as we increase the dimensionality. Moreover, the braid reflect approximate algorithm fails less and needs less iterations to find an input that proves a token is argmaxable. For example, for the bottom right two figures most tokens are shown to be argmaxable with 1 or 0 iterations.}
\label{fig:random-uniform}
\end{figure*}

\begin{figure*}[h]
\includegraphics[width=\textwidth]{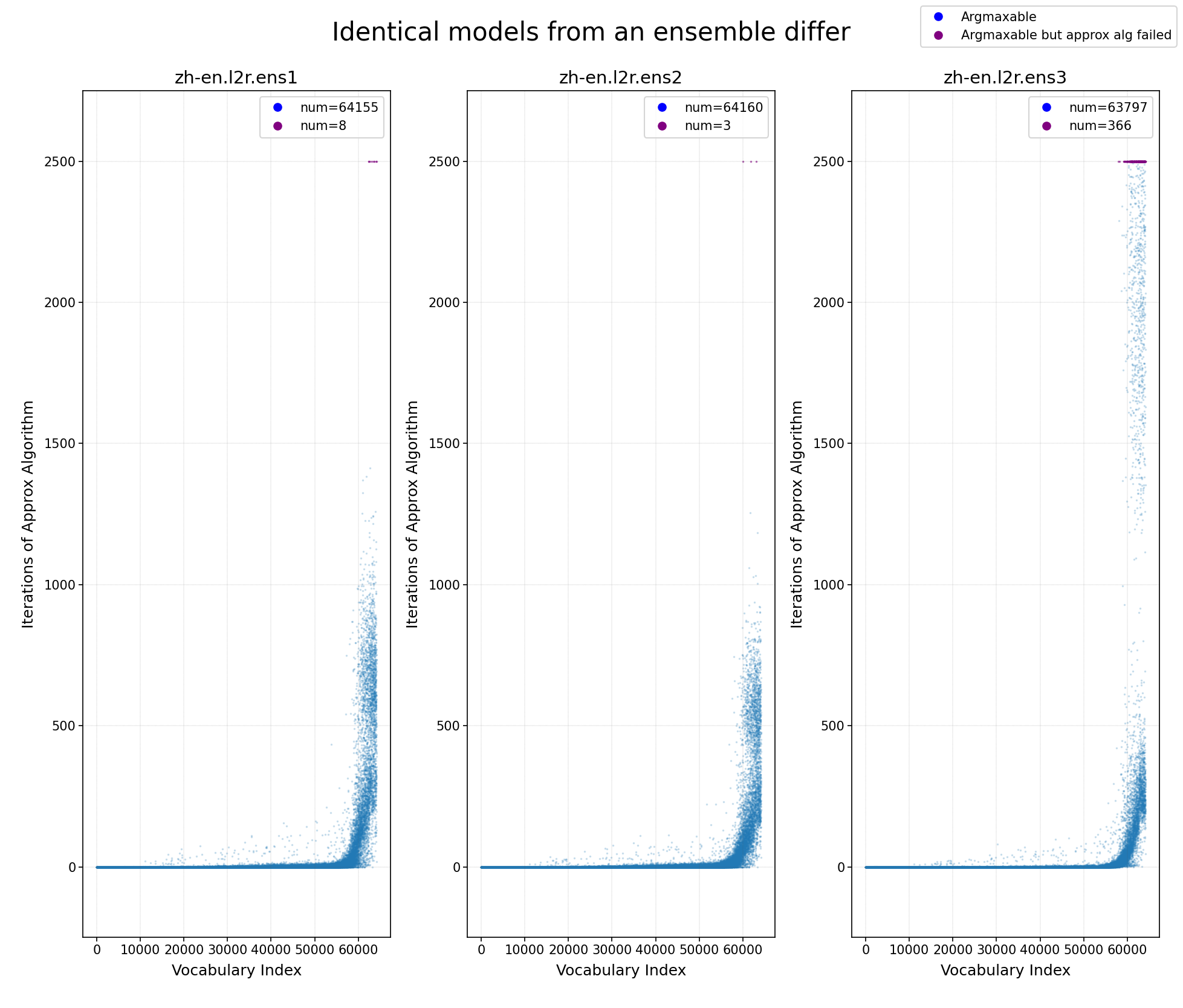}
\caption{Models from an ensemble can differ a lot in how easy they are to scan for unargmaxable tokens despite their difference being solely the random seed used in initialisation. As can be seen, the right-most figure has $366$ vocabulary tokens that are argmaxable but the approximate algorithm fails to find a solution, compared to $8$ and $3$ for the other two models.}
\label{fig:ensemble-diff}
\end{figure*}

\begin{figure*}[h]
    \begin{subfigure}[b]{0.95\textwidth}
        \includegraphics[width=\textwidth]{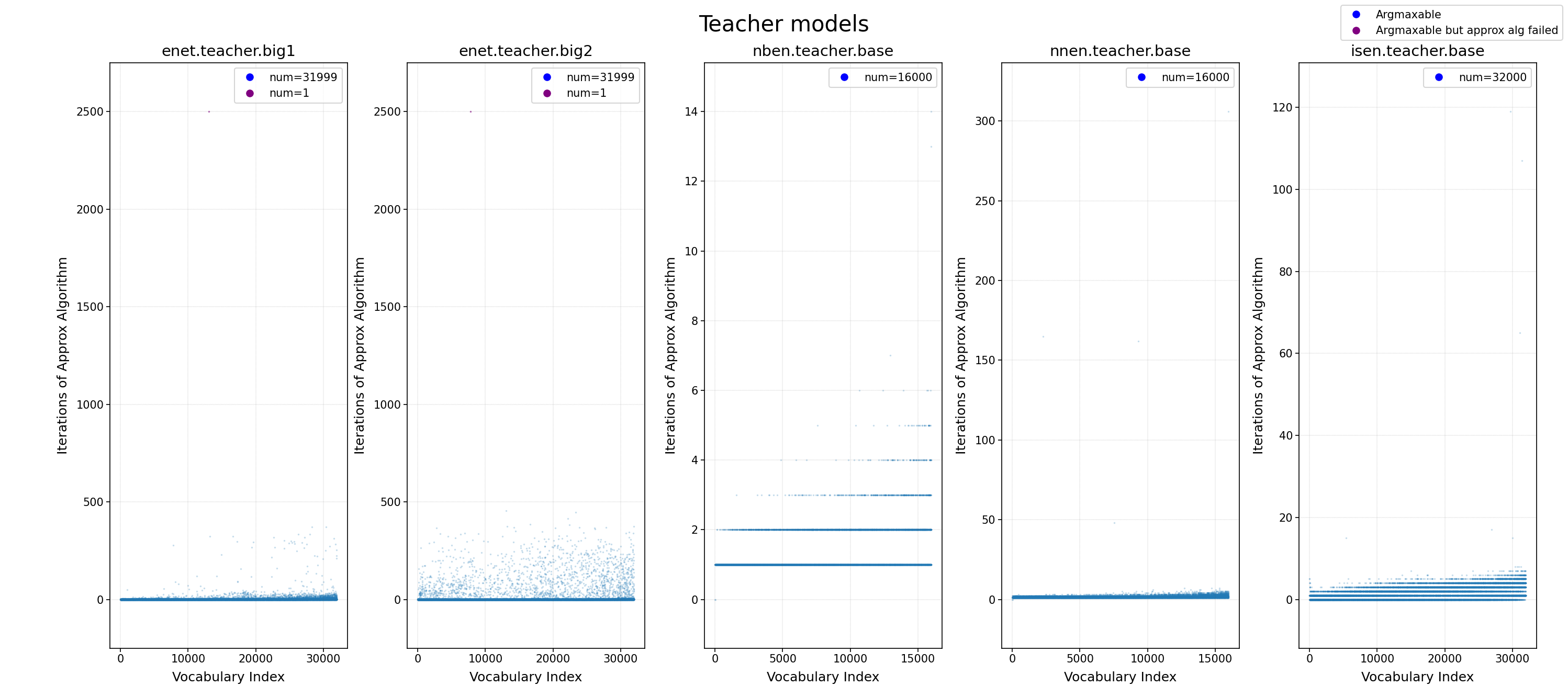}
        \caption{The approximate algorithm needs more iterations to show that the tokens of teacher models are argmaxable despite the dimensionality of the Softmax weights being larger than the student models.}
    \end{subfigure}
    \begin{subfigure}[b]{0.95\textwidth}
        \includegraphics[width=\textwidth]{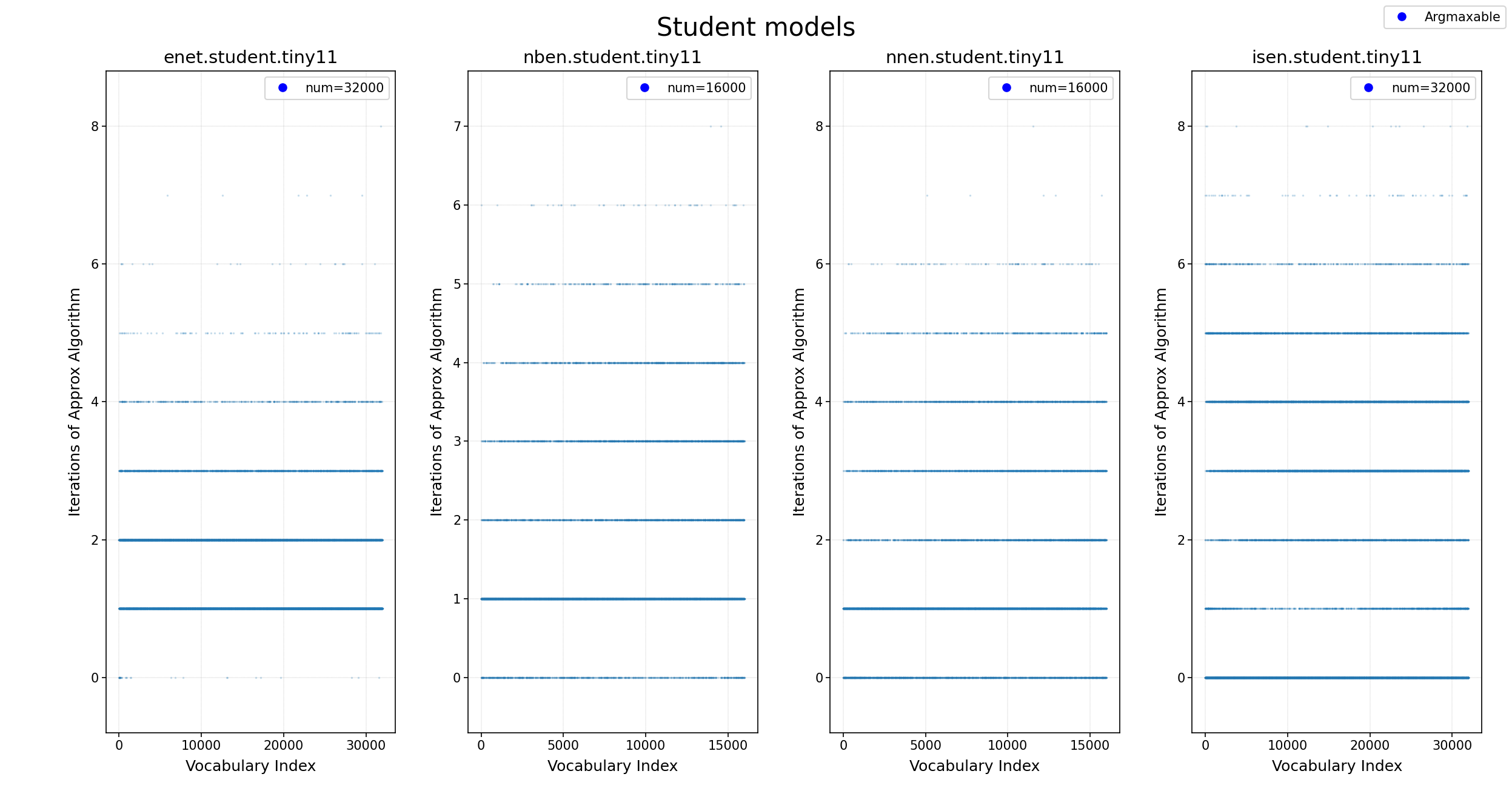}
        \caption{Student models can easily be shown to have argmaxable tokens.}
    \end{subfigure}
    \caption{Number of iterations of the approximate algorithm needed to show that a vocabulary token is argmaxable.}
\label{fig:teacher-student}
\end{figure*}

\FloatBarrier

\section{Activation Range of Softmax Layer Inputs}
\label{app:range}

Neural network activations are bounded in magnitude in practice, since larger activations can lead to larger gradients and instability during training. In this work, we made the assumption that the Softmax layer inputs $\vvec{x}$ are bounded within a range for all dimensions: $-100 \leq \vvec{x} \leq 100$. Below we provide some supporting empirical evidence that this assumption is reasonable.

We checked this assumption on 2 Helsinki NLP OPUS models for en-ru and bg-en, which were found to have unargmaxable tokens.
We took $10$ million sentence pairs from OPUS as released in ~\citet{tiedemann-2020-tatoeba} for the corresponding language pairs and input them to the corresponding models, decoding using the gold translations. We then recorded the range of the minimum and maximum activation for the Softmax layer inputs.

Since our assumption is that all $512$ dimensions are bounded between $-100$ and $100$, we focus on the range of the minimum and maximum activation for each output token across all dimensions. We therefore calculate a 99 percentile for the min and max activation per token across all dimensions as well as the overall min and max activations overall. The results can be seen in Table~\ref{tab:input-range}, from which we can see that for these two models our assumption holds for all activations produces for 10 million sentences and the percentiles show that more than 99\% of the extreme values fall within the $[-50, 50]$ range.

\begin{table}[h!]
\scalebox{0.75}{
\begin{tabular}{c | c c c c}
\toprule
model & min range & max range & min & max \\
\midrule
bg-en & $[-37.5, -9.4]$ & $[12.1, 40.3]$ & $-57.47$ & $58.87$ \\
en-ru & $[-41.6, -9.9]$ & $[10.9, 36.4]$ & $-95.4$ & $94.4$ \\
\bottomrule
\end{tabular}
}
\caption{Range of activations for Softmax inputs as calculated on $10$ million sentence pairs from OPUS. Ranges are 99 percentiles and min and max are the largest activation across all dimensions for all sentences.}
\label{tab:input-range}
\end{table}
\end{document}